\documentclass[11pt,onecolumn]{IEEEtran}
\usepackage[cmex10]{amsmath} 
\usepackage{graphicx,setspace}
\usepackage{fullpage}
\usepackage{subfig}

\usepackage{cite}

\title{
Multi-parameter acoustic imaging of uniform objects in inhomogeneous media
}
\author{
H.~Emre G{\"u}ven\thanks{H.~E.~G{\"u}ven is with the Department of Electrical and Computer
Engineering, Northeastern University, Boston, MA 02115.}, %
  Eric~L.~Miller\thanks{E.~L.~Miller is with the Department of Electrical and Computer
Engineering, Tufts University, Medford, MA 02155.}, %
  and~Robin~O.~Cleveland\thanks{R.~O.~Cleveland is with the Department of Mechanical
Engineering, Boston University, Boston, MA 02215.}
}
\date{\today}
\begin{document}

\doublespacing
\maketitle
\begin{abstract}
The problem studied in this paper is ultrasound image reconstruction from frequency-domain measurements of the scattered field from an object with contrast in attenuation and sound speed. The case where the object has uniform but unknown contrast in these properties relative to the background is considered.  Background clutter is taken into account in a physically realistic manner by considering an exact scattering model for randomly located small scatterers that vary in sound speed. The resulting statistical characteristics of the interference is incorporated into the imaging solution, which includes applying a total-variation minimization based approach where the relative effect of perturbation in sound speed to attenuation is included as a parameter. Convex optimization methods provide the basis for the reconstruction algorithm. Numerical data for inversion examples are generated by solving the discretized Lippman-Schwinger equation for the object and speckle-forming scatterers in the background. A statistical model based on the Born approximation is used for reconstruction of the object profile. Results are presented for a two dimensional problem in terms of classification performance and compared to minimum-$\ell^2$-norm reconstruction. Classification using the proposed method is shown to be robust down to a signal-to-clutter ratio of less than 1~dB.
\end{abstract}

\section{Introduction} \label{sec:intro}

	The last two decades have seen significant interest in the development and evaluation of the lesion formation process induced by High Intensity Focused Ultrasound (HIFU) for noninvasive cancer treatment \cite{bsk1,bsk2,bsk3,bsk4,bsk5,bsk6,bsk7,bsk8,bsk9,bsk10}.  HIFU treatment effectively increases the temperature within an intended region inside the tissue, thus forming a lesion that serves the purpose of eliminating cancerous cells non-invasively. Effective monitoring of the lesions is an essential requirement for the treatment to be successful. 
	
	The state of the art in imaging for HIFU treatment is achieved by tracking the temperature changes via magnetic resonance imaging.  A more convenient and less expensive modality for use in guiding HIFU treatment is ultrasound. Recent research in this area includes monitoring HIFU treatment via temperature tracking using two dimensional ultrasound \cite{ebbini}.  While some studies show that ultrasound backscatter may reveal information regarding the temperature distribution following HIFU exposure \cite{kaczkowski}, it is difficult to obtain ultrasound images showing the changes induced by the HIFU process in general, as the backscatter properties of the tissue do not change unless cavitation occurs, a process that is typically avoided \cite{hall,sanghvi}.
Physics-based approaches \cite{karbeyaz08,vandongen} have thus been examined to show the feasibility of HIFU monitoring by analyzing traditional ultrasound RF data using wave-based imaging techniques to identify the regions that have increased sound speed and acoustic attenuation.
	
	As a result of HIFU application, the sound speed of the affected tissue has a slight contrast (about 1\%) but the change is reversible and thus even if imaging is possible based on the sound speed changes \cite{simon,pernot}, the lesion may no longer be visible once the tissue cools. One property whose contrast changes significantly with a rise in temperature is the acoustic attenuation (80--700 \%), and the alteration is irreversible even after the tissue is cooled \cite{tyreus}. Therefore, an ultrasound imaging modality based on attenuation is potentially suitable to this application. 

	Acoustic tomographic imaging is a well-developed field where much of the early work \cite{devaney,kak} was based on the dual approach of sampling $k$-space in circular arcs.  In ultrasound imaging for biomedical applications there is typically a very limited-view of the region of interest which severely degrades the quality of images that can be obtained using diffraction tomography \cite{kak}. Some of the recent work in this area has concentrated on using physics-based models to facilitate the solution of the inverse problem \cite{EricPaper1,EricPaper2,karbeyaz08}. The use of the Born model in solving the imaging problem studied here is motivated by the simplicity it provides in modeling and computation. In the case of HIFU lesions, the use of the Born model is justified partially on physical grounds based on the relatively low contrast with a maximum of 1\% in sound speed and 1.7\% in the ratio $\alpha_p/k_b$ (the ratio of perturbation in attenuation to the background wavenumber) \cite{karbeyaz08}. While the large size of the lesions relative to the insonifying wavelengh does, strictly speaking, violate the Born model; the results of using the Born model to invert data consistent with the exact scattering physics will be shown in Sec.~\ref{sec:results} to be sufficiently accurate  in the context of a limited view inverse problem to provide suitable results.

	A significant challenge in many sensing contexts is the interference from unknown variations in the background that can overwhelm the signal from the object of interest. In the case of ultrasound, sub-resolution perturbations to the background properties manifest themselves as speckle in B-mode images. There has been significant effort in literature to characterize the formation of speckle using statistical methods. An accepted model for the signal magnitude due to coherent integration of the scattered field from numerous smaller scatterers is given by the Rayleigh distribution \cite{burckhardt,wagner}. Motivated by this work, a Rayleigh distributed model is used in this paper for the amplitude of the speckle forming background clutter where the location of the inhomogeneities come from a uniformly random distribution in space.

The use of total-variation regularization, which is well-known in the inverse problem literature \cite{rudin}, is especially successful when the signal of interest has piecewise smooth features.  Under a uniform lesion model \cite{zhou}, HIFU lesions are suitable for imaging with total-variation reconstruction because the binary (lesion vs background) classification of the region of interest naturally yields a piecewise-smooth structure.  
 In the approach taken here, total-variation reconstruction is used as the objective function of a convex optimization problem. 
The exact model employed in this paper is formed by spatially coincident contrast profiles in attenuation and sound speed \cite{karbeyaz08}. 
This model allows the characterization of the reduction of the number of parameters by a factor of two, due to the relationship between the two profiles. 

The reconstruction method is demonstrated using simulation data generated by solving the Lippman-Schwinger equation for the object and the background variation in a discrete setting. While linear addition of speckle forming signals \cite{jensen1,jensen2,liuffc} are useful for characterizing the distribution of texture in B-mode images (i.e., processed, time domain data), here a fundamentally different approach is taken in that the clutter is defined as the spatial distribution of the random, constitutive properties of the background medium \cite{burckhardt,wagner}.  This necessitates a first principles approach, described in Sec.~\ref{sec:theory},  where the effect of clutter is taken into account by solving the Lippman-Schwinger equation (LSE) in a discretized setting to generate the frequency-domain data, based on the spatial distribution of inhomogeneities in the background, as will be explained in Sec.~\ref{sec:results}. Solving the LSE enables a more realistic simulation of the data contributed by the lesion and background clutter compared to data generated using a matched, Born model. Thus, the solution of LSE is performed to include the region of interest where the lesion resides, in addition to all background scatterers that reside within the elliptical rings in associated with the time-of-flight for the corresponding transmitter-receiver pair. Thermal noise is simulated by linear addition of white Gaussian noise to the solution of Lippman-Schwinger equation. The reconstruction results are quantified using binary classification performance for the object vs background pixels. Lastly, the performance of the proposed method is compared with the minimum-$\ell^2$-norm reconstruction.

\section{Background} \label{sec:backgr}

\subsection{Problem statement}
	As motivated in Section~\ref{sec:intro}, in this paper we are interested in imaging an object that has contrast both in sound speed and attenuation with respect to the background.  
Data are obtained by means of a linear array of ultrasonic elements operated in multistatic mode.  That is, one element is used as a transmitter and then all elements (including the transmit element) are used as receivers to collect backscatter data.  The system then steps through until all elements have been used to transmit. Further details regarding the setting for numerical experiments are specified in Section~\ref{sec:results}.

  \subsection{Mathematical background}

	The partial differential equation describing the propagation of time-harmonic acoustic waves is the Helmholtz equation \cite{chew}. The Helmholtz equation associated with a constant-density medium is employed here per previous work \cite{karbeyaz08}:
	\begin{align} \label{eq:pde1}
		\nabla^2 \phi(r) + k^2(r) \phi(r) = - f(r) .
	\end{align}
where $\phi(r)$ is the total acoustic field, $k(r) = \omega / c(r)$ is the wavenumber associated with angular frequency $\omega$ and sound speed $c(r)$, and $f(r)$ is the source function.
For the special case that $k(r)=k$, a uniform background medium, the solution $g(r,r')$ to Eq.~\eqref{eq:pde1} for a point source $f(r) = \delta(r-r')$ is defined as the free space Green's function, and for 2-D space is given by $g(r,r') = (j/4) H_0^{(1)}(k|r-r'|)$ \cite{chew}. To find the total acoustic field in terms of the Green's function, expressing the space-dependent squared-wavenumber $k^2(r)$ as a sum of a constant background component $k_b^2$ and a scatter component $k_s^2(r)$, one arrives at %
	\begin{align}
		\left[ \nabla^2 + k_b^2 \right] \phi(r) = -f(r) - k_s^2(r) \phi(r) .
	\end{align}
	Consequently, $\phi(r)$ satisfies
	\begin{align}
		\phi(r) &= \int g(r,r') f(r') dr' + \int g(r,r') k_s^2(r') \phi(r') dr'  \label{eq:pre_LS} \\
			  &= \phi_b(r) + [\mathcal{G}_{s} \phi](r)
	\end{align}
	where the first integral term in \eqref{eq:pre_LS} is recognized as the background field $\phi_b(r)$ (the field that would be present if there were no inhomogeneities), and $\mathcal{G}_{s}$ represents the last term in Eq.~\eqref{eq:pre_LS} in an integral operator form that can be interpreted as a mapping from the total field to the scattered field due to $k_s^2(r)$. This results in the Lippman-Schwinger equation \cite{chew}:
	\begin{align} \label{eq:LS}
		\left[ \mathcal{I} - \mathcal{G}_{s} \right] \phi(r) = \phi_b(r)
	\end{align}
	where $\mathcal{I}$ is the identity operator, i.e., $[\mathcal{I} \phi] (r) = \phi(r)$. The scattered field can then be expressed as the difference between the total field and background field, so that
	\begin{align}
		\phi_s(r) &= \phi(r) - \phi_b(r) \\
		&= \left\{ \left[ \mathcal{I} - \mathcal{G}_{s} \right]^{-1} -  \mathcal{I} \right\}\phi_b(r) \label{eq:op_id}
	\end{align}
	When the scattering profile is weak, %
	the operator on the right hand side of Eq.~\eqref{eq:op_id} may be approximated by 
	\begin{align}
		\phi_s(r) \approx \mathcal{G}_{s} \phi_b(r) = \int g(r,r') k_s^2(r') \phi_b(r') dr'
	\end{align}
	that is known as the (first-order) Born approximation \cite{chew,kak}. 
	
	Let $l$, $m$ denote the numbers enumerating the elements of an ultrasound array whose elements are used to sequentially transmit pulses and simultaneously receive their returns, where $l$ denotes the transmitting element and $m$ denotes the receiving element. The measurements in the frequency domain corresponding to element pair $(l,m)$ are characterized by the Born model to relate physical properties of the object to scattered field measurements as formulated in\cite{karbeyaz08}, and are given by
	\begin{align} \label{eq:measurements}
		b_{l,m}(\omega) =\int_V k_s^2(r,\omega) h_l(r,\omega) h_m(r,\omega) d^3r + \nu_{l,m}(\omega)
	\end{align}
	such that $h_l(r,\omega)$ and $h_m(r,\omega)$ are the spatial transfer functions for transmitting element $l$ and receiving element $m$, found by integrating the corresponding Green's function over the element surface; and $\nu_{l,m}(\omega)$ is the associated thermal noise term. The scattering component of the wavenumber is given by \cite{karbeyaz08}
	\begin{align} \label{eq:ks2}
		k_s^2(r,\omega) =  - j \frac{2\omega}{c_b} \alpha_p(r,\omega) - \frac{2\omega^2}{c_b^3} c_p(r) - \frac{2\omega^2}{c_b^3} c_s(r)
	\end{align}
	where $c_p(r)$ and $\alpha_p(r,\omega)$ are the perturbation in sound speed and frequency-dependent attenuation profile of the object, respectively, and $c_s(r)$ is the background variation in sound speed that produces speckle.  For soft tissue the attenuation increases approximately linearly with frequency \cite{szabo} and therefore the  frequency-dependent attenuation term can be expressed as $\alpha_p(r,\omega) = \psi_p(r) \cdot \omega / (2\pi)$, where $\psi_p(r)$ is expressed in units of Np/(Hz$\cdot$m) \cite{karbeyaz08}. 

We now invoke the assumption that the object of interest, that is the HIFU lesion, is uniform and both the sound speed and attenuation have the same spatial distribution.   The ratio of the second term on the right hand side of Eq.~\eqref{eq:ks2} to the first term is thus a constant within the object:
         \begin{align} \label{eq:mudef}
         	\mu %
			&= \frac{2\pi \, c_p(r)}{c_b^2 \, \psi_p(r)} .
         \end{align} 
The scattering contribution can therefore be expressed as:
	\begin{align}
		k_s^2(r,\omega) =-\frac{j\omega^2}{\pi c_b} \left[ \psi_p(r) - j \mu \psi_p(r) \right] - \frac{2\omega^2}{c_b^3} c_s(r).
	\end{align}

	The equations expressed above in integral forms are typically discretized and solved using digital computers, which requires expressing the variables and data in discrete space. In order to achieve a faithful characterization of the Born integral in Eq.~\eqref{eq:measurements}, a sufficient spatial sampling rate is essential. The integral representing the Born model Eq.~\eqref{eq:measurements} used for the reconstruction is discretized in space to form a matrix-vector representation of the integral. The rows of the matrix $A$ correspond to the pointwise multiplication of the spatial transfer functions such that the entry $a_{\iota,\kappa}$ at row-$\iota$ and column-$\kappa$ of the matrix $A$ is given by
	\begin{align}
		a_{\iota,\kappa} &= -\frac{j\omega_\iota^2\Delta}{\pi c_b} h_{l_\iota}(r_\kappa,\omega_\iota) h_{m_\iota}(r_\kappa,\omega_\iota) 
	\end{align} 
	where each different (transmitter, receiver, frequency) combination $(l_\iota,m_\iota,\omega_\iota)$ is assigned a different row $\iota = 1, \ldots, M$; column-$\kappa$ of $A$ corresponds to a different pixel location $r_\kappa$ for $\kappa = 1, \ldots, N$ covering the region of interest where the lesion resides; and $\Delta$ is the grid area. Let $\lambda_{min}$ denote the wavelength corresponding to the highest frequency at which measurements are obtained. A spatial sampling rate of 8 samples per wavelength ($\lambda_{min}$) was employed in order to achieve acceptable characterization of the integral. The frequency domain measurement $b_\iota$ due to the lesion, under the Born approximation, can now be expressed as
	\begin{align}
		b_\iota = (1 - j \mu ) \sum\limits_{\kappa=1}^N a_{\iota,\kappa} \psi_p(r_\kappa) + n_\iota
	\end{align}
	where $n_\iota$ is the total interference originating from two sources, i.e., the scattering from background inhomogeneity and the thermal noise $\nu_\iota$ at the receiver:
	\begin{align}
		n_\iota & =\int  \frac{-2\omega_\iota^2}{c_b^3} c_s(r) h_{l_\iota}(r,\omega_\iota) h_{m_\iota}(r,\omega_\iota)d^2r +\nu_\iota . \label{eq:newborn}
	\end{align}
 Hence Eq.~\eqref{eq:measurements} can be approximated in discrete space by a matrix-vector product 
	\begin{align} \label{eq:measure}
		b = (1 - j \mu ) A x + n
	\end{align}
	where $x = [ \psi_p(r_1) \ldots \psi_p(r_N) ]^T$ is the vector of contrast values in attenuation depending on the pixel location; $b = [b_1 \ldots b_M]^T$ is the vector of data points; and $n = [n_1 \ldots n_N]^T$ is the discrete noise vector studied in further detail in the next section. The problem to be addressed is the estimation of $x$ and $\mu$ from $b$.

  \subsection{Statistical modeling of speckle and thermal noise}

	The two main sources of noise in the data considered here are the sound speed variations in the background and the receiver noise. The receiver (thermal) noise is well characterized by white Gaussian distribution, however the characterization of the inhomogeneous background is more complex. Here, for purposes of inversion, a statistical approach is used for characterizing the inhomogeneous background, where a linear contribution of the background sound speed variation is used via the Born model \cite{jensen1,cramblitt} for deriving the statistics of the total noise in the measurements while maintaining mathematical tractability.  %
	
	The two features characterizing the inhomogeneities in the background are the location and the strength of each small scatterer.  The physical model for each scatterer is based on the accumulation of sub-pixel changes in the background properties.  A Rayleigh density is used to model the resulting amplitude distribution of such scattering \cite{burckhardt,wagner}.  While there is no exact model in literature replicating the contrast in sound speed of sub-resolution scatterers known to the authors at the time of writing, the Rayleigh distribution was used to model the sound speed increase due to small scatterers in the background.  Data in Sec.~\ref{sec:results} are simulated by first generating the location of each background scatterer randomly, and then setting the amplitude of the perturbation from a Rayleigh probability density function, which is parametrically controlled to yield varying signal-to-clutter ratio (SCR) levels. %
	
	Lastly, again based on the linear model, the total noise contribution from numerous scatterers weighted by the corresponding spatial transfer functions of the transmitting and receiving elements displays a normal distribution, which is numerically verified in the next subsection.  Therefore, a weighted squared-error minimization is appropriate due to its maximum likelihood estimator properties in the linear model \cite{scharf}, where the weighting matrix $W$ is given by the inverse square-root of the noise covariance matrix $C_n$.  In what follows, the first two moments (mean $m_n$ and covariance $C_n$) of the noise are analytically derived in terms of the parameters for the medium and the transducer elements, by using the Born  approximation, where $m_n$ and $W = C_n^{-1/2}$ are to be used in the reconstruction in the next section.

	In order to study the statistics of the noise $n$, Eq.~\eqref{eq:newborn} is analyzed for a model with point scatterers with a uniform random spatial distribution in the region of interest, the sound speed variation profile can be represented by a sum of delta functions 
	\begin{align} 
		c_s(r) & =\sum\limits _{i=1}^{N_s}s_{i}\,\delta(r-r_{i})
	\end{align}
	where $N_s$ is the number of scatterers, $s_{i}$ is the random perturbation value with a Rayleigh distribution for the $i$-th scatterer \cite{cramblitt}, and $r_{i}$ is the corresponding random location. 	 
	By substituting the above expansion into Eq.~\eqref{eq:newborn} and changing the order of summation and integration, the noise can be expressed as
	\begin{align}
		n_\iota %
		 & =\sum\limits _{i=1}^{N_s} \frac{-2\omega_\iota^2 }{c_b^3} h_{l_\iota,\omega_\iota}(r_{i})h_{m_\iota,\omega_\iota}(r_{i}) s_i +\nu_\iota \\
		 & =\sum\limits _{i=1}^{N_s} K_{\iota}(r_{i}) s_{i}+\nu_\iota\label{eq:model1}
	 \end{align}
	where 
	\begin{align}
		K_\iota(r_{i}) = \frac{-2\omega_\iota^2}{c_b^3} h_{l_\iota,\omega_\iota}(r_{i})h_{m_\iota,\omega_\iota}(r_{i})
	\end{align}
	denotes the Born kernel \cite{karbeyazPhd} for transmitter-receiver pair $l_\iota$-$m_\iota$, frequency $\omega_\iota$. %
	Furthermore, by vectorizing the variables in Eq.~\eqref{eq:model1} for transducer element pairs $(l,m)$, and frequency $\omega$, the notation can be simplified as
	\begin{align}
		n &= \left( \sum\limits _{i=1}^{N_s} K(r_i) \cdot s_i \right) +\nu \label{eq:model3}
	\end{align}
	such that $n = [n_1 \ldots n_M]^T$, $\nu = [ \nu_1 \ldots \nu_M ]^T$, and $K(r_i) = [ K_1(r_i) \ldots K_M(r_i) ]^T$.

    \subsection{Calculation of noise statistics}

It has been shown that fully developed speckle results in a Rayleigh distributed amplitude \cite{cramblitt}. While in the model above individual amplitudes of the scatterers in the background are Rayleigh distributed, the summation over a large number of sound speed contrast values multiplied by their corresponding weights $K(r_i)$ result in a jointly normal interference in the measurement vector. This observation is consistent with the Rayleigh envelope of the received signal from speckle \cite{burckhardt,wagner} due to the relation between circular Gaussian and Rayleigh distributions. In order to verify the joint normality of the observation vector, 1000 sets of randomly located scatterers were simulated with Rayleigh amplitude for the example setting explained in Sec.~\ref{sec:results}, and the joint normality of the observation vector was verified using Royston's multivariate normality test \cite{roys1} implemented in Matlab \cite{roysMatlab}. The $p$-value associated with the Royston test (the probability of obtaining a test statistic at least as extreme as the one observed, given that the null hypothesis is true, i.e., that the generating distribution is normal in this case) was found to be 0.0864.  For a significance of 0.05, the null hypothesis is not rejected since $p>0.05$, thus justifying the use of normal distribution in basing the reconstruction method on the use of mean and covariance as described in the next section.
	
	Following the physical model derived above for random scatterers in the background, the mean $m_n$ and covariance $C_n$ of the noise vector $n$ can be found by
	\begin{align}
		m_{n} &= \sum\limits_{i=1}^{N_s} E[ K(r_i)  s_i ]+E[\nu] \\
		&= N_s \cdot E[ s_i ] \cdot E[ K(r_i) ]  \\
		&= N_s \cdot \sigma_s \sqrt{\pi/2} \cdot \int\limits_{r_i\in R_s} \frac{1}{\alpha_{s}} K(r_i) d^2r_i  \label{eq:mean_n}
	\end{align}					
where we have assumed $\nu$ is zero mean, $\alpha_s$ denotes the area of the region of integration $R_s$. The covariance matrix is computed as
	\begin{align}
		C_{n} & =E[nn^{\ast}] - m_n m_n^{\ast} \\
		&= \sum\limits_{i=1}^{N_s} E[ K(r_i)  s_i s_i^{*}K^{*}(r_i) ]+E[\nu\nu^{*}] - m_n m_n^{\ast} \\
		& =\sum\limits_{i=1}^{N_s} E[s_i s_i^*] E\left[ K(r_i) K^{*}(r_i) \right] +C_{\nu} - m_n m_n^{\ast} 
	\end{align}
where we used the fact that $K s$ and $n$ belong to independent noise sources and hence are uncorrelated; and that the scattering strength $s_i$ is Rayleigh distributed where $s_{i}$ and $r_{i}$ are uncorrelated, with
	\begin{align}
		E\left[s_{i} s_{j}^*\right] & =2\sigma_{s}^{2}\cdot \delta(i-j)
		\label{eq:cov_s}
	\end{align}
	where %
	$\sigma_s$ is the mode of the Rayleigh probability distribution.

\begin{figure}
\centering
\includegraphics[scale=0.42]{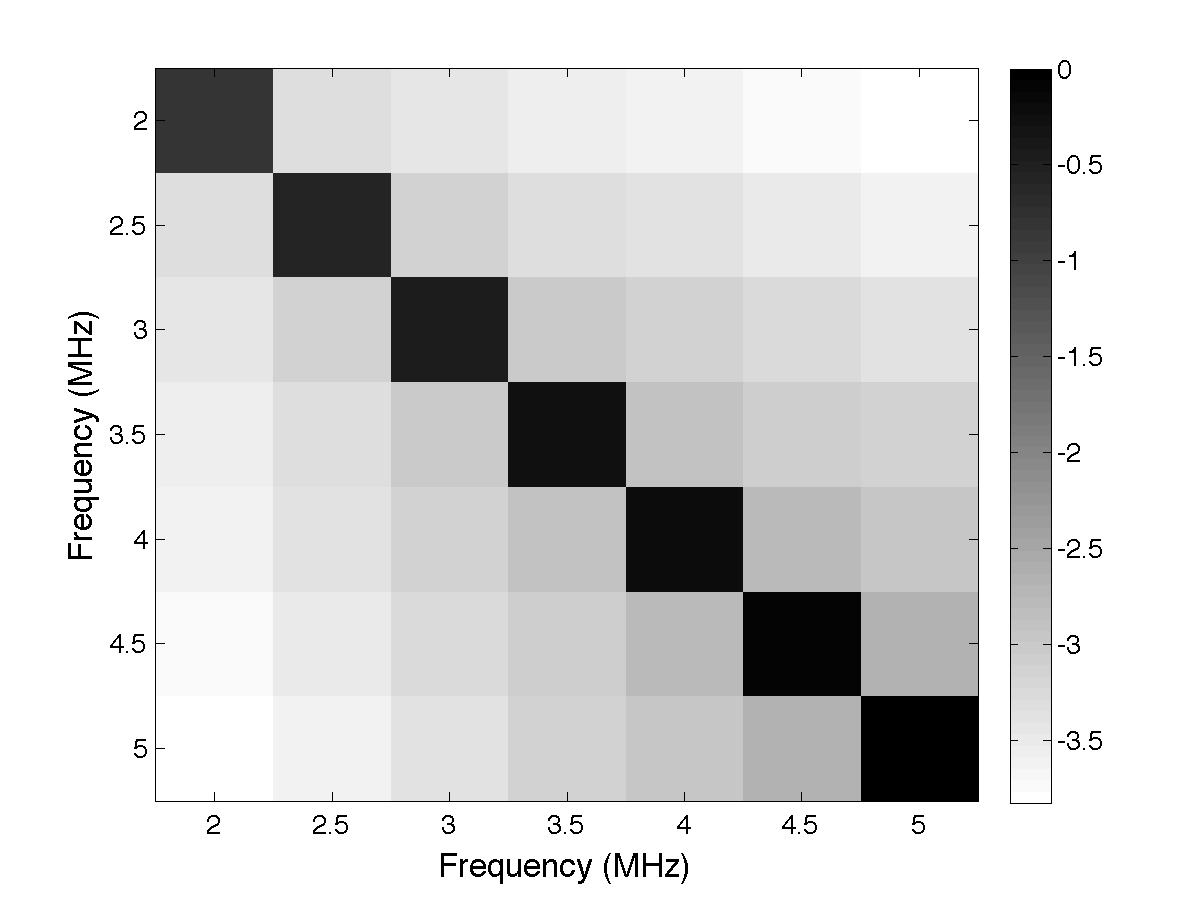}
\caption{Covariance sub-matrix corresponding to the use of middle element in both transmission and reception at different frequencies. Base-10 logarithm of the absolute value is shown for each entry.} \label{fig:cov_midEl}
\end{figure}

If the additive Gaussian noise component  is taken to be white, so that $C_\nu = \sigma_\nu^2 I$, then the noise covariance matrix is given by:  
	\begin{align}
		C_{n} & = 2 \sigma_{s}^{2} \sum\limits_{i=1}^{N_s} E[ K(r_i) K^{*}(r_i)]+C_\nu - m_n m_n^{\ast} , \\
		&= 2 \sigma_s^2 N_s \int\limits_{r_i\in R_s}  \frac{1}{\alpha_{s}}  K(r_i) K^{*}(r_i) d^2r_i +  \sigma_{\nu}^{2}I - m_n m_n^{\ast} . \label{eq:cov_n}
	\end{align}
The integral in Eq.~\eqref{eq:cov_n} can then be approximated by summation over a discrete grid on $r_i \in R_s$. For computing the covariance for the examples in the next section,  a spatial sampling rate of 8 samples per wavelength ($\lambda_{min}$) is used in both dimensions. Figure~\ref{fig:cov_midEl} shows the sub-matrix of $C_n$ computed for the region of interest used in the examples in Sec.~\ref{sec:results} corresponding with the use of the middle element of the array in both transmission and reception, as a function of frequency, normalized by the largest diagonal value of the sub-matrix for $\sigma_\nu=0$ and $\sigma_s = 1$.

\section{Framework for estimation} \label{sec:theory}

	This section provides the technical framework used to estimate the object profile from frequency domain measurements of the scattered field. %
	The whitening of the data is directly incorporated into the optimization problem by including the error-weighting (whitening) matrix $W = C^{-1/2}_n$ for reconstructing the original image $x$ from the measurement vector $b$. Therefore we aim to find the set of parameters $x$, $\mu$ that best represents the data as specified by the constraints of the problem, expressed by
	\begin{equation}
		\begin{array}{cc}
		\underset{x, \mu} {\mbox{minimize}} &  TV\left[\left(1-j\mu\right)x\right] \\ 
		\mbox{subject to} 
		& \Vert W [(1 - j \mu) A x  - b] \Vert_{2}  \leq \epsilon \\
		& x \geq 0
	\end{array}\label{eq:epsmuwhite_global}
	\end{equation}
where $TV[(1-j \mu) x]$ is the total variation (1-norm of the gradient) \cite{rudin} of the contrast profile $x$ multiplied by $(1-j\mu)$, $\mu$ is the scalar defined in Eq.~\eqref{eq:mudef}, and $\epsilon$ is the error radius allowed in the reconstruction. 
	    	
	The optimality of weighted minimum mean-squared error estimation for measurements with additive Gaussian noise as the maximum likelihood estimator is well-known \cite{scharf}. Generalized Tikhonov regularization is an example where a matrix-weighting (whitening) is applied to the error term \cite{vauhkonen}. While a constrained minimization form \cite{candesNoisy} is used here, the use of whitening for mean squared error minimization is justified through the equivalence of our formulation to that where the error in the predicted data and total variation are linearly combined to form the cost function to be minimized \cite{rudin}. Piecewise-constant functions tend to have lower total variation as the changes in such functions are sparser, and seeking a lower total variation thus favors such solutions \cite{rudin,candesNoisy}. Therefore a total variation based reconstruction is appropriate for HIFU lesions \cite{vandongen}, while the contribution of thermal noise and speckle is accounted for by the error radius allowed in the measurements. %
	
	The choice of regularization parameters for total variation reconstruction has been the subject of literature based on the relation between the total variation and the error in the observations \cite{rudin,vogel,wen,strong,candesNoisy}. According to the discrepancy principle \cite{morozov}, the error radius is chosen proportional to the expected noise level. While the optimal selection of the regularization parameter in the constrained formulation is beyond the scope of the current paper, a consistent method is used in setting the parameter $\epsilon$ in the examples of the next section. For the imaging of HIFU lesions studied here, an estimate for the total noise level in the data can be obtained by insonifying the medium prior to the formation of the lesion. %
Specifically, a realization of prior measurements $b_s$ from speckle-forming background variation and thermal noise is used to obtain $\epsilon = \Vert W b_s \Vert_2 / 2$ used in the reconstruction.  

Due to the multiplication of $x$ and $\mu$, \eqref{eq:epsmuwhite_global} is not convex in its variables. Nevertheless, for fixed values of $\mu$, the problem is convex in $x$. Consequently, \eqref{eq:epsmuwhite_global} is solved for a fixed $\mu$. The process is repeated for a range of discrete values $\mu = \mu_1,\dots, \mu_M$ and the solution for which the cost function $J(\mu,x) = TV[ (1-j\mu) x ]$ is minimum is chosen.
The performance of the overall estimation is discussed in the next section, where the contrast profile is used in a binary classification of the lesion vs background pixels. 

\section{Results} \label{sec:results}
     
In order to obtain faithful simulation of the physical principles of acoustic scattering, the Lippman-Schwinger equation is solved for the full scattering profile including the object and the background scatterers that result in speckle. The frequency domain data from the ultrasound elements are simulated by solving Eq.~\eqref{eq:LS} on a discrete grid with a spatial sampling interval of $\lambda_{min}/8$. The discrete Lippman-Schwinger equation is solved numerically using the Matlab implementation of the GMRES algorithm, which iteratively solves a linear equation to yield a minimum residual over the Krylov subspace \cite{saad}. %
The parameters used in calling the function \texttt{gmres} are the maximum number of iterations set to 500 and the tolerance in approximation set to $10^{-6}$.

\begin{figure}
\centering
\subfloat[][]{\includegraphics[scale=0.35]{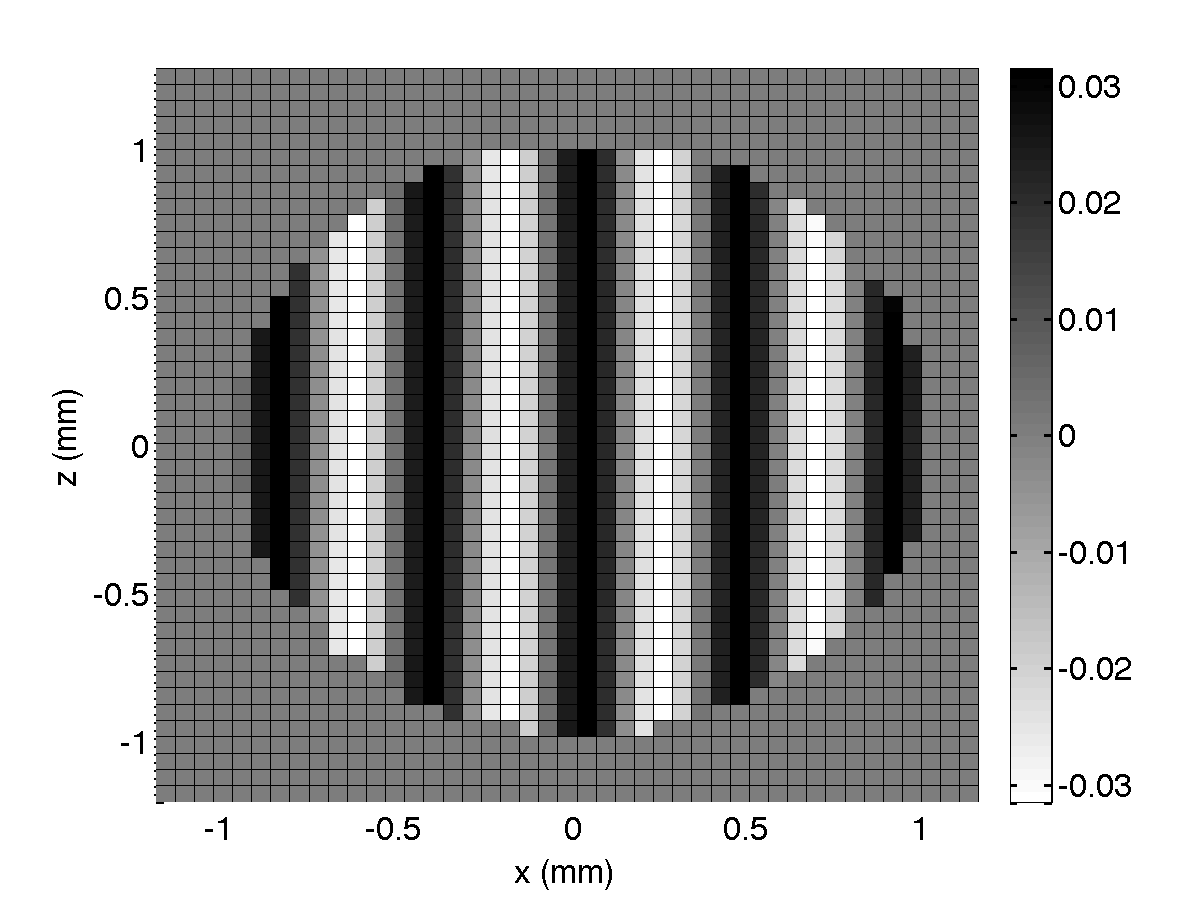}\label{fig:longley}}
\subfloat[][]{\includegraphics[scale=0.35]{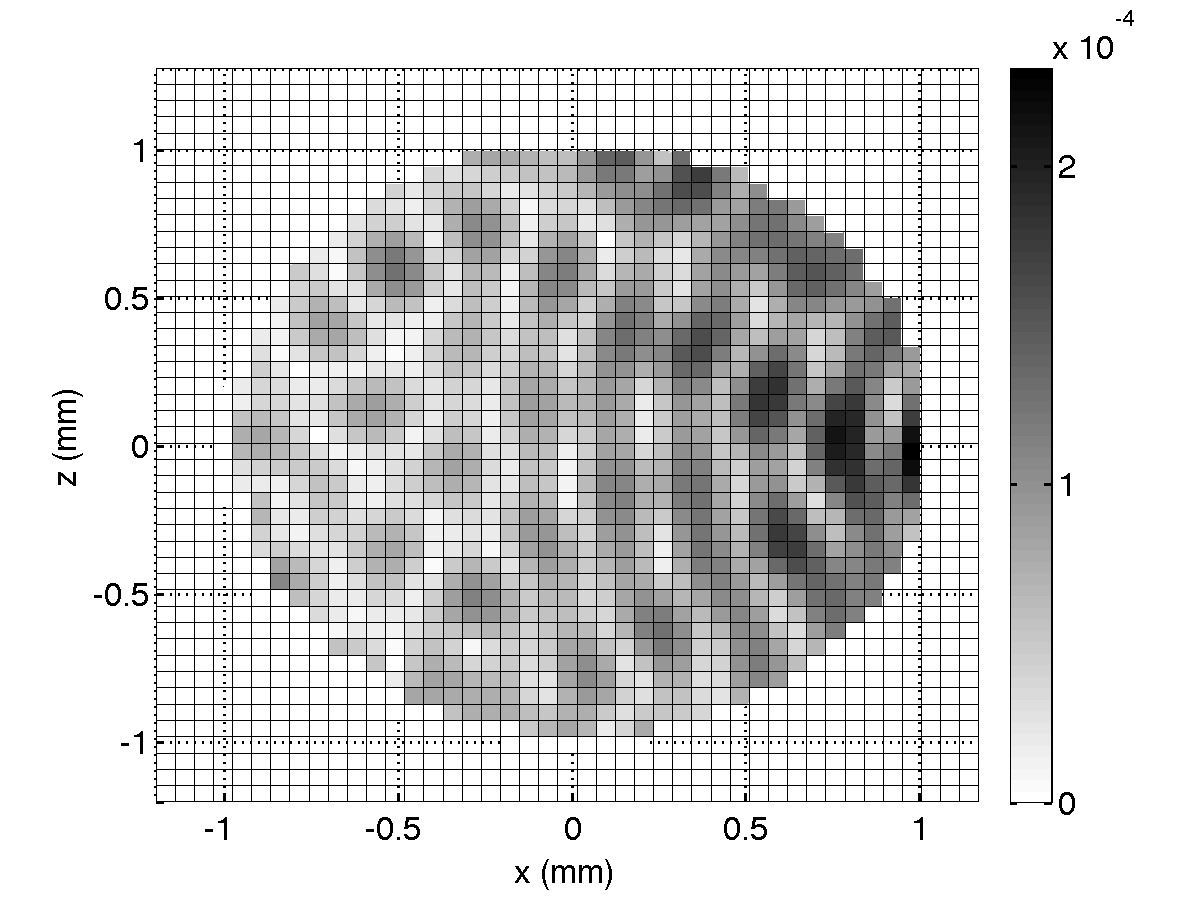}\label{fig:longleyErr}}
\caption{\subref{fig:longley} Real part of the total pressure field computed using the analytic solution given for the object interior in the case of an incident plane wave originating from far left \cite{longley}. \subref{fig:longleyErr} Magnitude of the difference between the analytical solution \cite{longley} and the numerical solution method employed in this paper.}
\end{figure}

In order to verify the validity of the solution in discrete space, the pressure field is compared to an analytic solution given for the interior of a circular object in the form of a series expansion for the case of an incident plane wave \cite{longley} for the case of an object with 2 mm diameter and 10 m/s contrast in sound speed from a background sound speed of 1540 m/s.  The result from the discretized Lippman-Schwinger equation $\phi_{LS}$ and the analytic solution $\phi_0$ are shown in Fig.~\ref{fig:longley}. The mean squared error (the relative norm of the difference between the two solutions $\Vert \phi_{LS} - \phi_0 \Vert_2 / \Vert \phi_0 \Vert_2$) is less than 0.27 percent. The magnitude of the difference between two calculated fields, i.e., $\vert \phi_{LS} - \phi_0 \vert$, is shown in Fig.~\ref{fig:longleyErr}. 

\begin{figure}
		\centering
		\includegraphics[scale=.8]{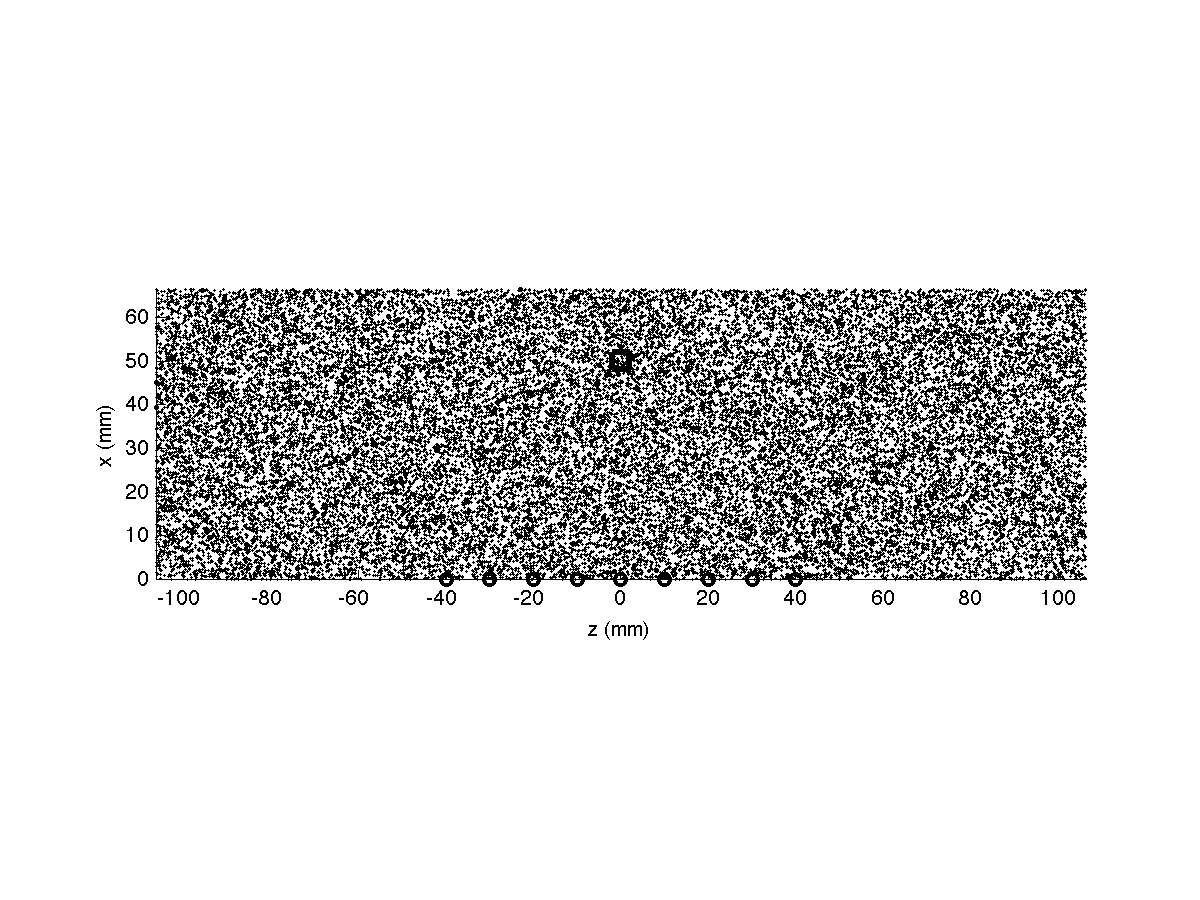}%
		\caption{Background variation generated as point scatterers within the shown rectangular region, where the object is located within the  $4\times4$ mm box centered at $z=0$, $x=50$ mm, and the transmitting-receiving elements are marked with circles on the $x=0$ line. } \label{fig:setting} 
	\end{figure}

Figure~\ref{fig:setting} shows the setting used to generate data for the reconstruction for a medium with both a target and background scatterers. The data are generated by solving Eq.~\eqref{eq:LS}, the Lippman-Schwinger equation, for the total contrast profile $k_s^2(r)$ that includes the lesion and the background scatterers as defined in Eq.~\eqref{eq:ks2}.  Data are generated directly in the frequency domain within 2-5 MHz with a step size of 500 kHz for each transmitter-receiver combination of all 9 elements of a linear array of point sources with inter-element distance of 10 mm. Typically there is significant prior information in monitoring the HIFU therapy due to the controlled nature of the lesion formation \cite{karbeyaz08}, and thus the region of interest can be adjusted according to the available prior information regarding the expected location of the lesion. The grid is placed at a 4 mm $\times$ 4 mm region around the object.  In practice, time-domain systems are used to collect data and a time-window is applied in order to reduce interference from regions that are not of interest at the data collection stage.  In order to incorporate the effect of time-windowing, we include the background scatterers that reside within two spatial ellipses defined by the corresponding time-window for each transmitting and receiving element pair, based on the time-of-flight and the size of the region of interest.  A more detailed explanation on the time-space domain relation is provided in the appendix, along with the time-window parameters used in the simulation.  The perturbation outside the grid is accounted for by setting the solution of Eq.~\eqref{eq:LS} to include all pixels that fall within the corresponding timing window and have a nonzero contrast relative to the background. The object used in this study is an ellipse of dimensions $2\times 2.4$ mm centered 50 mm away from the middle element of the array. This size is on the smaller range of HIFU lesions \cite{seip,ziadloo} but keeps the problem numerically tractable. The contrast in the sound speed inside the ellipse is 10 m/s relative to the background sound speed of 1540 m/s, and the contrast in attenuation is 10 Np/(MHz~m) relative to the lossless background. The background variation is generated in the form of uniformly located point scatterers.  

The number of background scatterers are set equal to the multiplication of the area of the covered region by a density of 250 scatterers per cm$^2$ similar to \cite{liuffc} (where a volumetric density of 4000 per cm$^3$ is assumed, i.e. $4000^{2/3} \approx 250$).  %
The locations of the scatterers are set by generating uniformly random coordinates in $x$ and $z$ in the area shown in Fig.~\ref{fig:setting}. 
Each scatterer of one pixel size (the same pixel size used both in the solution of Eq.~\eqref{eq:LS} on a discrete grid and also the reconstruction) is given a random contrast in sound speed with Rayleigh probability density functions with varying modes to obtain different levels of signal-to-clutter ratio (SCR).  The SCRs are calculated as $10 \log_{10}\left( \Vert b_l \Vert^2 / \Vert b_s \Vert^2 \right)$ from the  vectors $b_l$ and $b_s$ that are associated with the measurements due to the presence of lesion-only and speckle-only, respectively. For three different modes of Rayleigh amplitude for scatterers, the SCRs are 10.8, 5.2, and 0.3 dB. %
The corresponding values computed for $\epsilon = \Vert W b_s \Vert_2 / 2$ from $b_s$, as described previously, are 0.113, 0.215, and 0.307 (normalized by the norm $\Vert b \Vert_2$ of the total signal $b$ in the presence of lesion and speckle together), respectively. %
Following the solution of the Lippman-Schwinger equation for the field scattered from the lesion and the background scatterers, thermal noise is added to the resulting scattered field measurements where the noise variance is chosen to yield 30 dB SNR in the scattered field measurements.  

\begin{figure}
		\centering
		\includegraphics[scale=0.4]{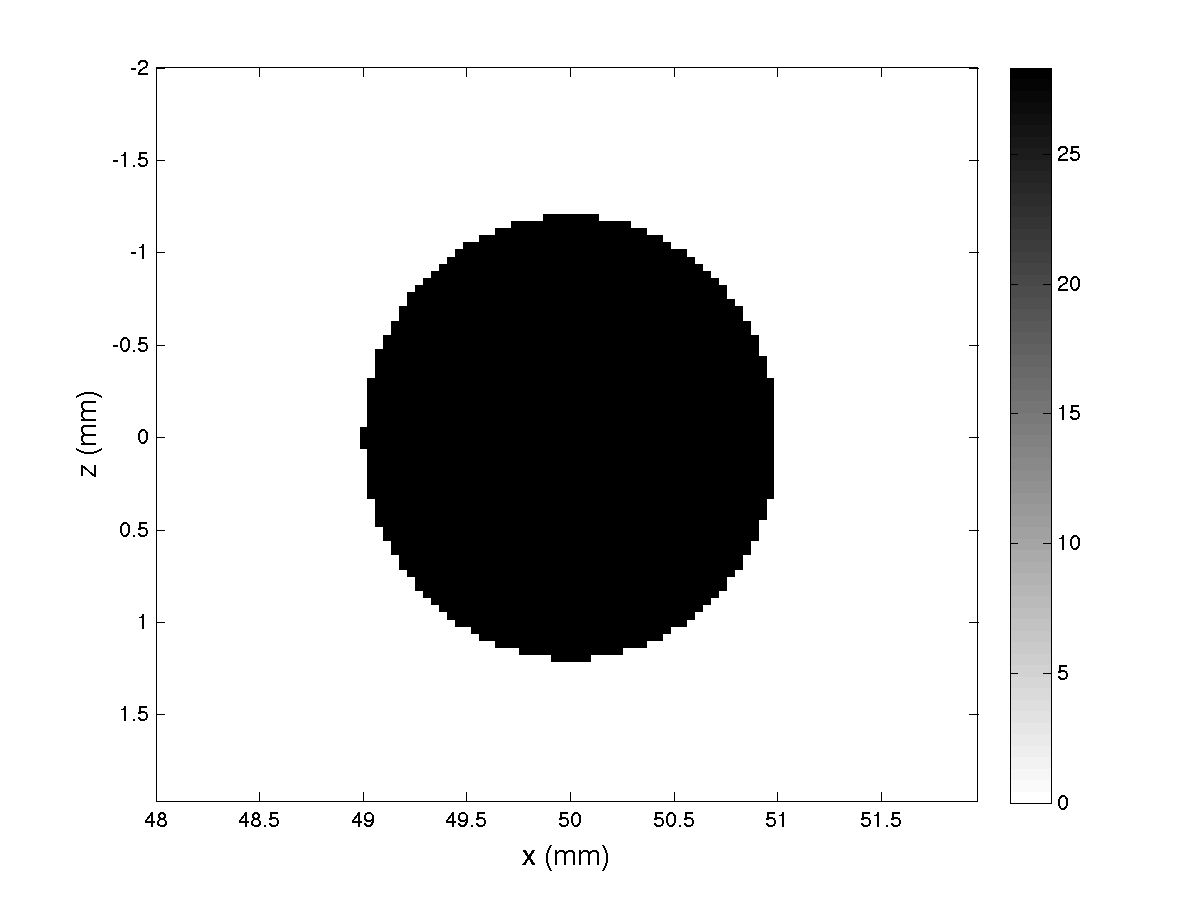}
		\caption{Contrast of the true object in attenuation in units of Np/(m MHz), multiplied by $|1-j\mu|$.}
		\label{fig:TVMorig} 
	\end{figure}

Figure~\ref{fig:TVMorig} shows the contrast profile $\vert1-j\mu\vert x$ only due to the presence of the lesion. %
The true value of $\mu$ is $(2\pi \cdot10) / (1540^2 \cdot 10^{-5}) \approx 2.65$, while the estimation is performed for a discrete set of candidate $\mu$ values on a grid with 0.1 interval between [0.5,10], hence used in the estimation of $x$ per \eqref{eq:epsmuwhite_global}.

\begin{figure}
  \centering
  \subfloat[10.8 dB SCR]{\includegraphics[scale=0.42]{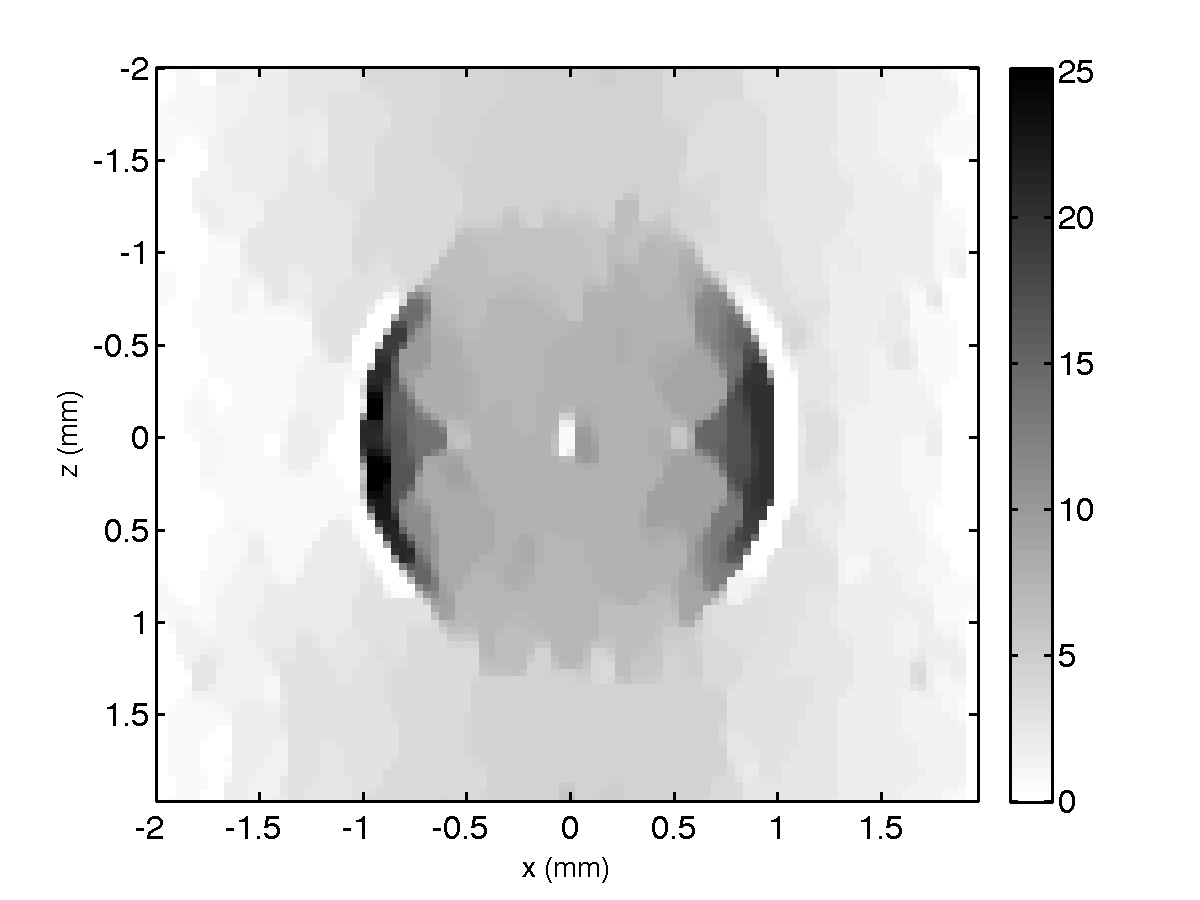} \label{fig:recon1}}
  \subfloat[Binarized reconstruction and borders of the actual lesion]{\includegraphics[scale=0.42]{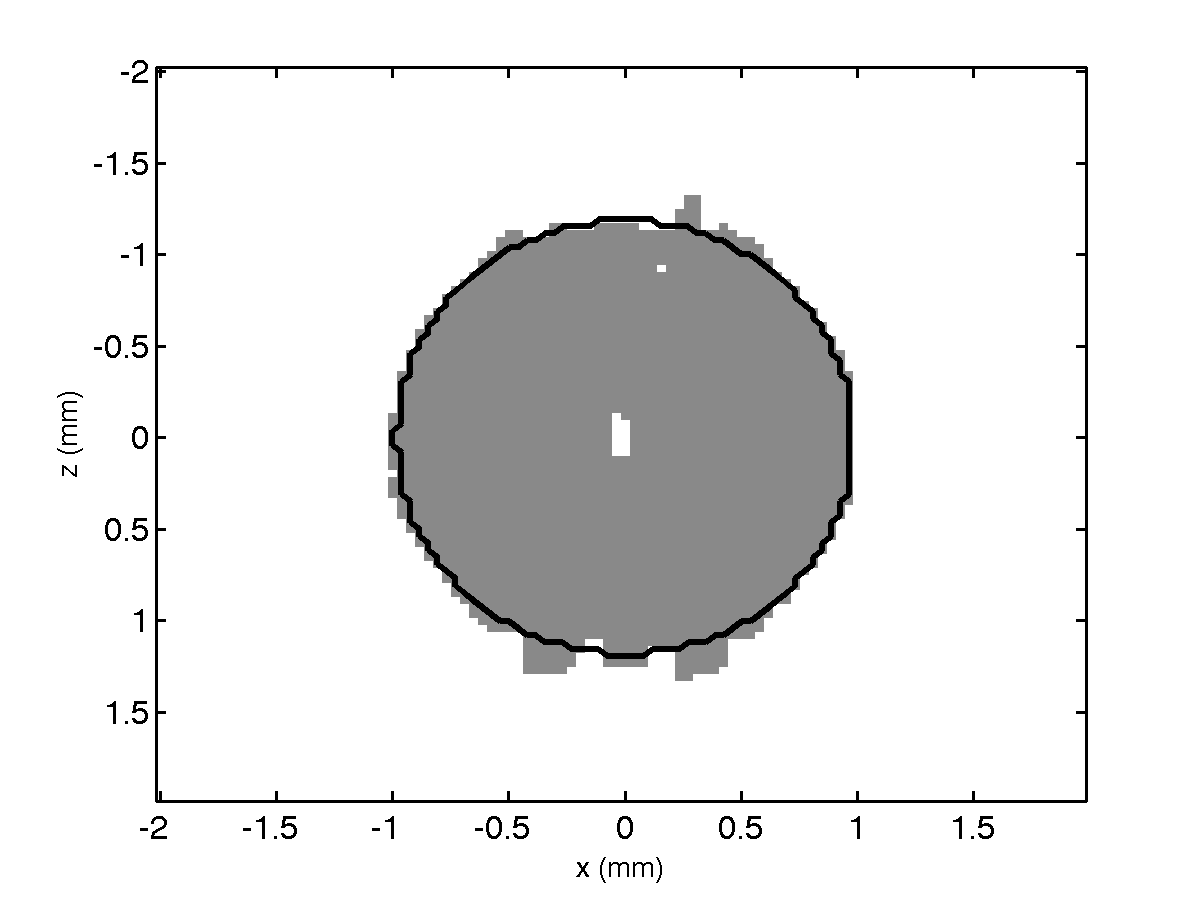} \label{fig:bin1}}
  \caption{\subref{fig:recon1} Contrast of the estimated object in attenuation in units of Np/(m MHz), multiplied by the magnitude $|1-j\hat{\mu}|$  for each case; \subref{fig:bin1} binarized (lesion vs non-lesion) reconstruction where the edges of the actual lesion are outlined with the solid black line.}
\end{figure}

\begin{figure}
		\centering
		\includegraphics[scale=0.42]{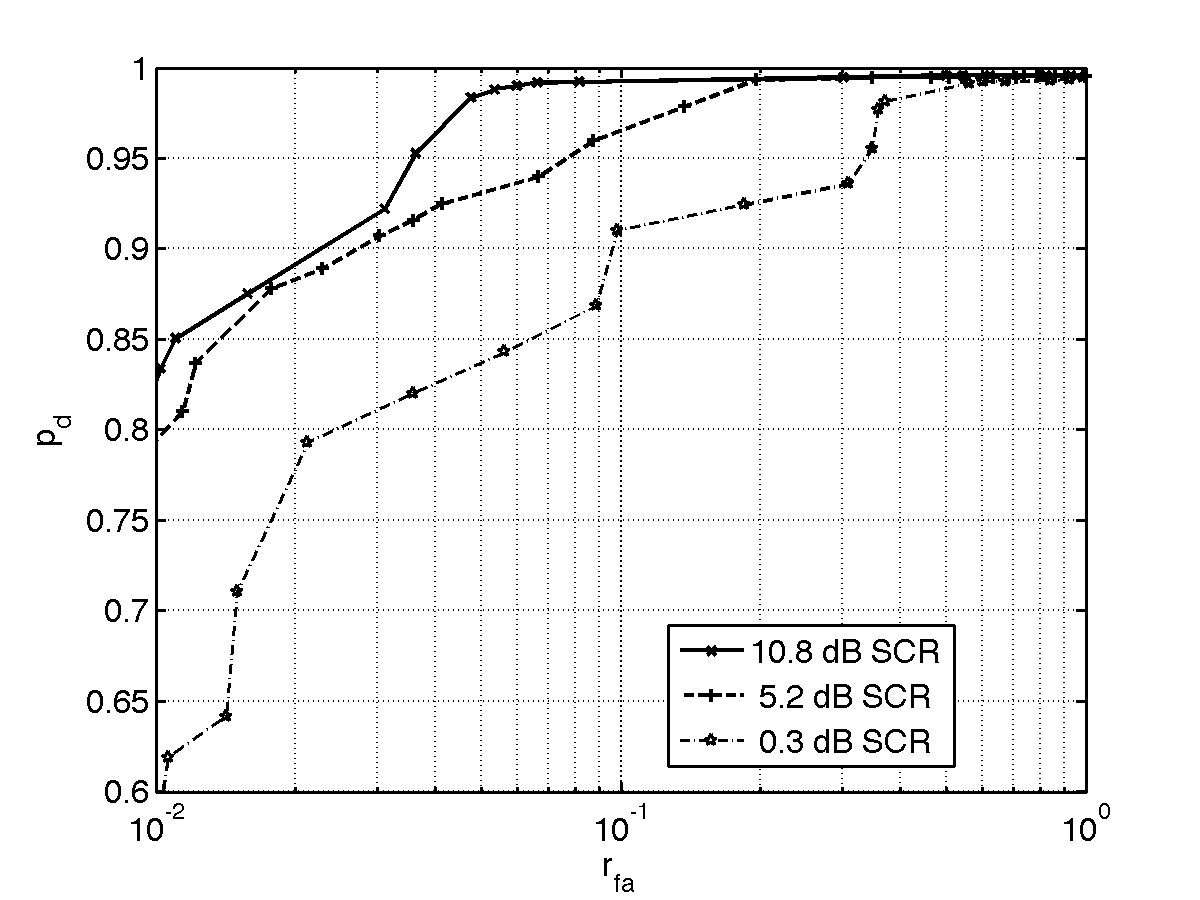}
		\caption{Binary classification performance curves (detection probability vs relative false alarm) for SCRs of 10.8, 5.2, and 0.3 dB SCR}
		\label{fig:PdPfa} 
	\end{figure}

Figures~\ref{fig:recon1}-\ref{fig:bin1} show the reconstruction results for the case with 10.8 dB SCR. In order to quantify the performance of the reconstruction, a binary quantization is employed for the classification of the lesion and non-lesion pixels in the reconstructed image. For different values of the threshold, the number of pixels correctly identified as lesion $n_{CL}$, and those that are misclassified as lesion $n_{ML}$ are counted. Both counts are divided by the number of true lesion pixels $n_{TL}$; resulting in an estimated probability of detection $p_d = 1-n_{CL}/n_{TL}$ for lesion pixels, and a relative false alarm rate $r_{fa}=n_{ML}/n_{TL}$, where %
the normalization for $r_{fa}$ is not a probability but expresses the number of the falsely classified background pixels relative to the lesion size. For example $r_{fa} = 0.05$ corresponds to the case where the number of misclassified background pixels is 1/20 of the total number of pixels in the true lesion. The performance curves for the cases with different SCRs are plotted in Fig.~\ref{fig:PdPfa}, obtained by setting the threshold as integer multiples of 1/100 of the peak value, shown in the interval $10^{-2} \leq r_{fa} \leq 1$. For a relative false alarm rate of $r_{fa}=0.05$, the probability of detection is near 0.98, 0.93, and 0.84 for the cases with 10.8, 5.2, and 0.3 dB SCR, respectively. The estimates $\hat{\mu}$ were 1.3, 1.2, and 1.1, respectively, compared to the actual value of 2.65. 

\begin{figure}
  \centering
  \subfloat[Reconstruction of object with 1/10 the contrast for 11.8 dB SCR]{\includegraphics[scale=0.42]{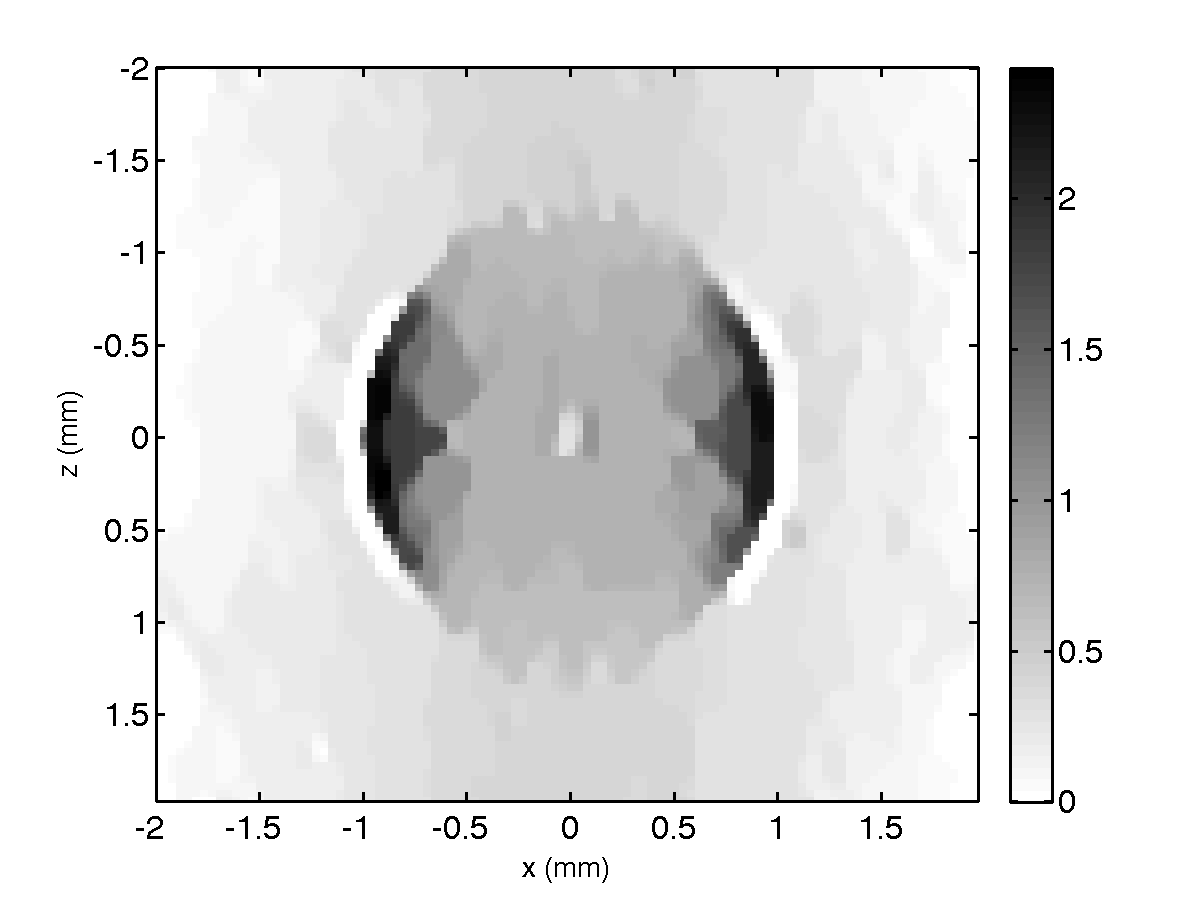} \label{fig:recon_a1c1}}
  \subfloat[Binarized reconstruction and borders of the actual lesion]{\includegraphics[scale=0.42]{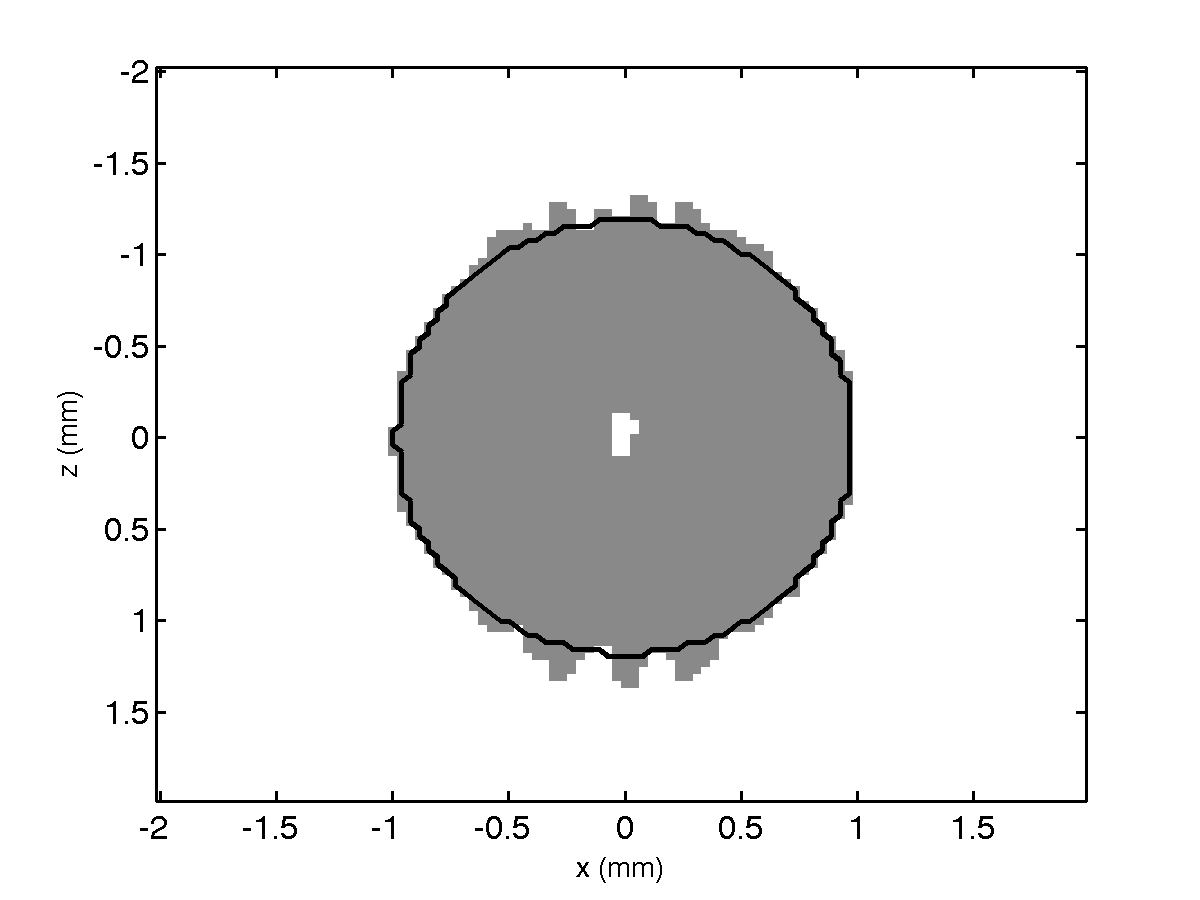} \label{fig:bin_a1c1}}
  \caption{\subref{fig:recon1} Contrast of the estimated object in attenuation in units of Np/(m MHz), multiplied by the magnitude $|1-j\hat{\mu}|$  for the object contrasting 1 m/s and 1 Np/(m MHz) in sound speed and attenuation, respectively; \subref{fig:bin_a1c1} binarized (lesion vs non-lesion) reconstruction and the edges of the actual lesion.} \label{fig:recons_and_binary}
\end{figure}

\begin{figure}
		\centering
		\includegraphics[scale=0.42]{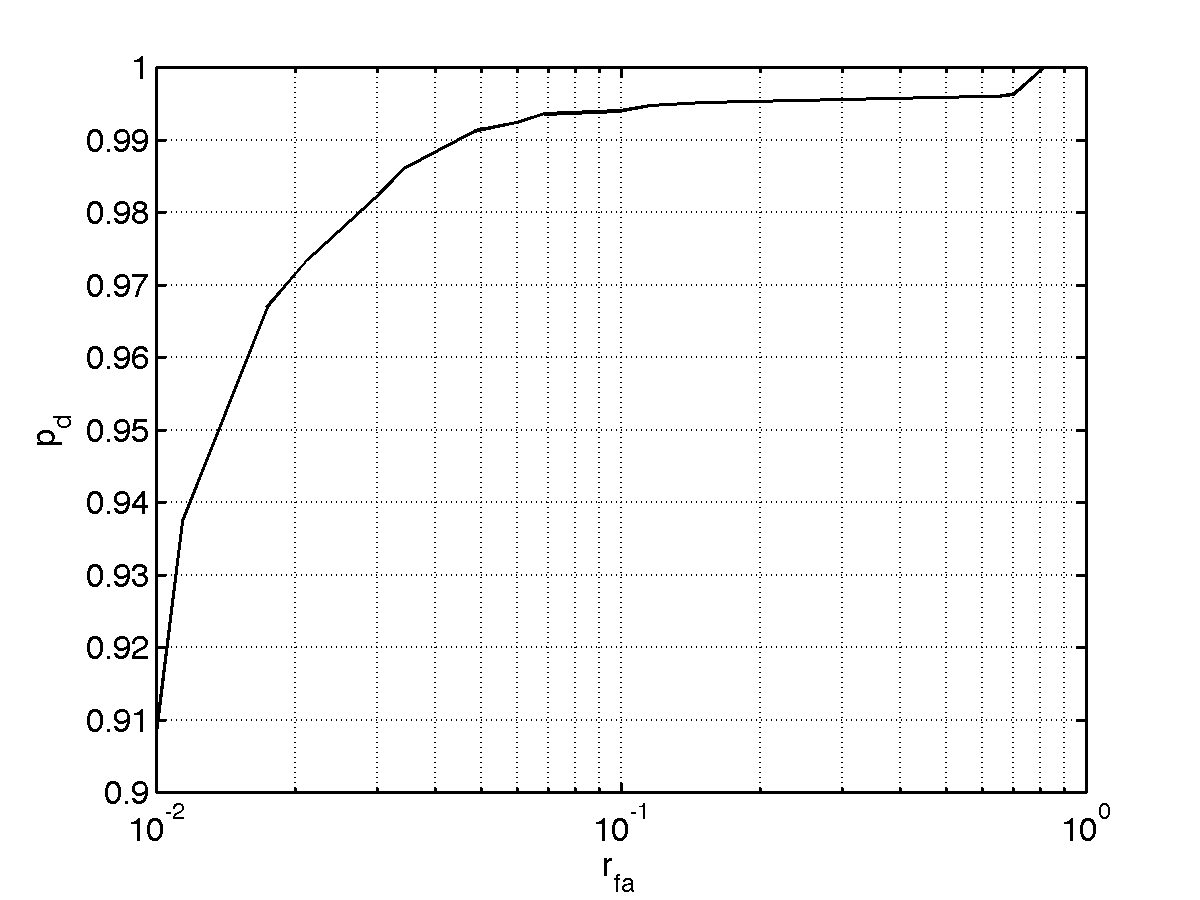}
		\caption{Binary classification performance curve for lesion with 1/10 of the contrast and 11.8 dB SCR.}
		\label{fig:PdPfa_a1c1} 
	\end{figure}

To validate that estimates of $\mu$ improve in cases with lower interference, the reconstruction is performed for the same sized object with lower contrast in sound speed (1 m/s) and attenuation (1 Np/(m MHz)) for an SCR of 11.8 dB, where the error in the Born model is smaller due to the lower contrast relative to the background. Figures~\ref{fig:recons_and_binary}~and~\ref{fig:PdPfa_a1c1} show the associated reconstruction results and the classification performance. In this instance, the estimate of $\mu$ is 2.3, significantly closer to the true value 2.65, and the detection probability was $p_d \approx 0.99$ for $r_{fa}=0.05$. Therefore, while the estimation of $\mu$ is affected by the total interference in the measurements (model error plus noise), the classification performance curves in Fig.~\ref{fig:PdPfa} indicate that the ability to estimate the boundary of the lesion is robust to estimation errors in $\hat{\mu}$. 
  
\begin{figure}
  \centering
  \subfloat[Min-$\ell^2$-norm reconstruction for 10.8 dB SCR]{\includegraphics[scale=0.42]{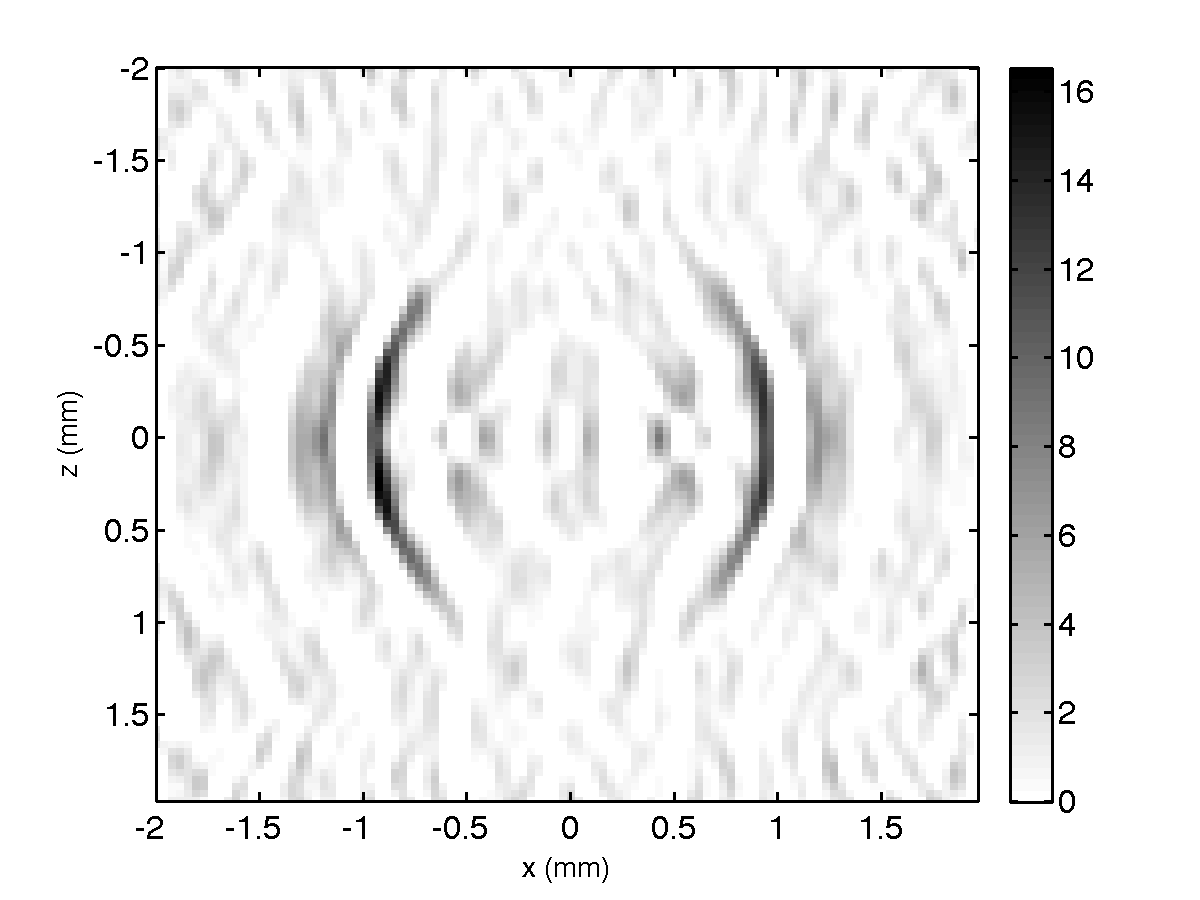} \label{fig:ell2_recon1}}
  \subfloat[Binary classification performance]{\includegraphics[scale=0.42]{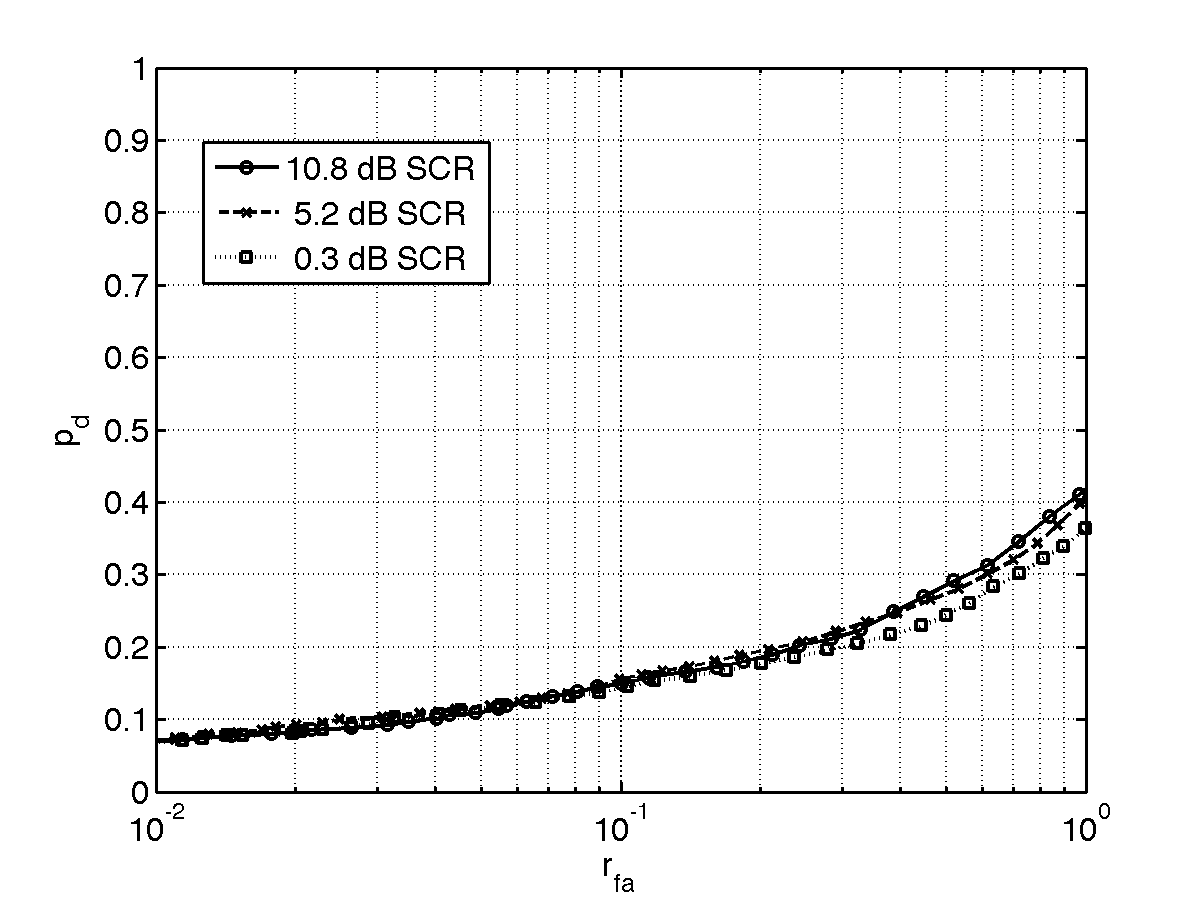} \label{fig:PdPfa_ell2}}
  \caption{\subref{fig:ell2_recon1} Contrast of the object in attenuation for minimum-$\ell^2$-norm estimation, multiplied by the magnitude $|1-j\mu|$  at 10.8 dB SCR; \subref{fig:PdPfa_ell2}  Detection probability vs relative false alarm for the three cases.}
\end{figure}

Lastly, we compare the performance of the proposed method to the case where the $\ell^2$-norm of the variable $x$ is used as the objective function instead of total variation. The use of $\ell^2$-norm as a penalty function added to the residual is well-known as Tikhonov regularization \cite{vauhkonen,vogel}, which improves the conditioning of the problem to enable a numerical solution. In order to maintain a similar setting in comparison to the application of our method, an optimization form, similar to \eqref{eq:epsmuwhite_global} is used, except that $\mu$ is assumed known for the objective function and data constraint. Specifically, the following optimization problem is solved
	\begin{equation}
		\begin{array}{cc}
		\underset{x} {\mbox{minimize}} &  \Vert x \Vert_2 \\ 
		\mbox{subject to} 
		& \Vert W [(1 - j \mu) A x  - b] \Vert_{2}  \leq \epsilon \\
		& x \geq 0
	\end{array}\label{eq:epswhite_ell2}
	\end{equation}
where the true value $\mu = 2.65$ is prescribed, and all other parameters are the same as in \eqref{eq:epsmuwhite_global}. Figures~\ref{fig:ell2_recon1} shows that, despite the use of the true value of $\mu$, the minimum-$\ell^2$-norm method is not able to reconstruct the object as well as the TV algorithm: the object is not dark inside and it is surrounded by ripples. For comparison purposes, Fig.~\ref{fig:PdPfa_ell2} shows the resulting classification performance from these reconstructions. %
The performance curve is significantly worse with a failure of the classification $p_d<0.13$ for $r_{fa}=0.05$ compared to $p_d\approx0.98$, $p_d\approx0.94$, and $p_d\approx0.81$, for SCRs of 10.8, 5.2, and 0.3, respectively, as seen in Fig.~\ref{fig:PdPfa} for the TV based method.
In comparison, the performance of the TV based is even better when the true value $\mu=2.65$ is prescribed, resulting in $p_d\approx0.99$, $p_d\approx0.95$, and $p_d\approx0.83$, for SCRs of 10.8, 5.2, and 0.3, respectively.

\section{Conclusions and future work} \label{sec:conclude}

	In this paper a method was presented to image a lesion with uniform sound speed and attenuation profile in 2-D. The methods used in this paper can be extended to problems in three-dimensional case with sufficient engineering effort for solving the convex optimization problems in higher dimensions. The method used for reconstruction is based on the Born model for frequency-domain measurements of the backscattered ultrasound signal from a linear array of elements.  By relating the scattered field measurements to the contrast profiles of the object in sound speed and attenuation, where one profile is expressed as a scalar times the other, the dimension of the estimation problem is reduced by half. 
  
	A model for the speckle forming clutter was employed using a random distribution of point sources where the scattering amplitudes were chosen so that the statistics matched a Rayleigh distribution. The Born approximation was then used to derive the statistics of the total noise in the observations. In order to take advantage of the structure of the interference due to the inhomogeneous background consisting of randomly located small scatterers, we performed a covariance analysis and applied whitening to the measurement model.
  
	For the numerical experiments, frequency-domain data were generated by solving the Lippman-Schwinger for the lesion, where the background scatterers were also included in the solution of the Lippman-Schwinger equation. The validity of the numerical solution to Lippman-Schwinger equation was checked by comparison to analytical expressions for a case previously studied in literature. Thermal noise in the measurements was simulated by linearly adding zero-mean white Gaussian noise, while speckle generating clutter was accounted for directly as perturbation in sound speed in the solution of Lippman-Schwinger equation. 
	
	For solving the problems \eqref{eq:epsmuwhite_global} and \eqref{eq:epswhite_ell2} we used \texttt{CVX}, a package for specifying and solving convex programs \cite{cvx,dcp}. For larger problems with higher computational load, it is possible to implement problem-specific solutions to the convex optimization problem \eqref{eq:epsmuwhite_global} to avoid the overhead of using general purpose convex solvers \cite{vogel,mattingley}.

	Our study of the classification performance shows the feasibility of the reconstruction method down to a signal-to-clutter ratio of less than 1~dB for identifying the lesion and background pixels. The detection rate of the TV based reconstruction method generally exceeded 80\%. In comparison, a minimum-$\ell^2$-norm reconstruction under similar constraints even when the true value of the scalar parameter is assumed to be known, had a detection rate of less than 15\%. These results suggest that TV based reconstruction is inherently more suited to the piecewise smooth nature of the HIFU lesions, as well as that a TV approach to detecting HIFU lesions can make detection of lesions by ultrasound practical.

\appendix[Time windowing of measurements] \label{app:timespace} 
As mentioned in the referring text above, time-domain systems are used in practice in order to obtain ultrasound measurements. Time windows are typically applied to avoid large signal returns from nearby scatterers. In our example application, the expected region for the object (HIFU lesion) location is known by design, and therefore application of a time-window is useful for avoiding excess interference that are due to background scatterers outside the region of interest.

While simulating the application of a time-window involves solving the LSE for the entire domain of scatterers for all frequencies of interest, finding the corresponding time-domain sequence by inverse Fourier transformation, and then applying a time-window to the resulting sequence; application of a spatial window involves a Born approximation for scatterers that are outside a close time-proximity of the object of interest by ignoring the secondary interactions between such scatterers. 
Due to the small size and scattering strength of the sub-resolution scatterers used in our numerical experiments, we adopt this approach in the simulations to make the simulation feasible.  Note that all secondary interactions within the spatial region associated with the time-window for a given element pair are accounted for in the solution of Eq.~\eqref{eq:LS}, including those of the lesion pixels and the background scatterers, both between themselves and each other. Below arguments are thus applicable under these conditions. 

For a given medium with approximately constant sound speed, the time-of-flight of a pulse is determined by the distance it travels in space. For a point element, the scattering location for a pulse emitted at $t=0$ and returning at $t = t_1$ resides on a circle with radius $r_1 = c \cdot t_1/2$ where $c$ denotes the speed of sound. Therefore, in a homogeneous background, a time-window applied between $[t_1,t_2]$ is associated with an annular region whose distance to the point element remains between $[r_1,r_2]$, where $r_2 = c \cdot t_2/2$. For the examples above, the spatial window at a distance between $[47, 53]$ mm from the center element corresponds to a time-window of $[61.04, 68.83]$ $\mu$s for $c = 1540$ m/s. For a pulse with 3.5 MHz center frequency and 3 MHz bandwidth, the duration is on the order of 0.3 $\mu$s and thus a time-window width of $T=7.8$ $\mu$s covers an interval of sufficient width for processing.

\begin{figure}
		\centering
		\includegraphics[scale=.6]{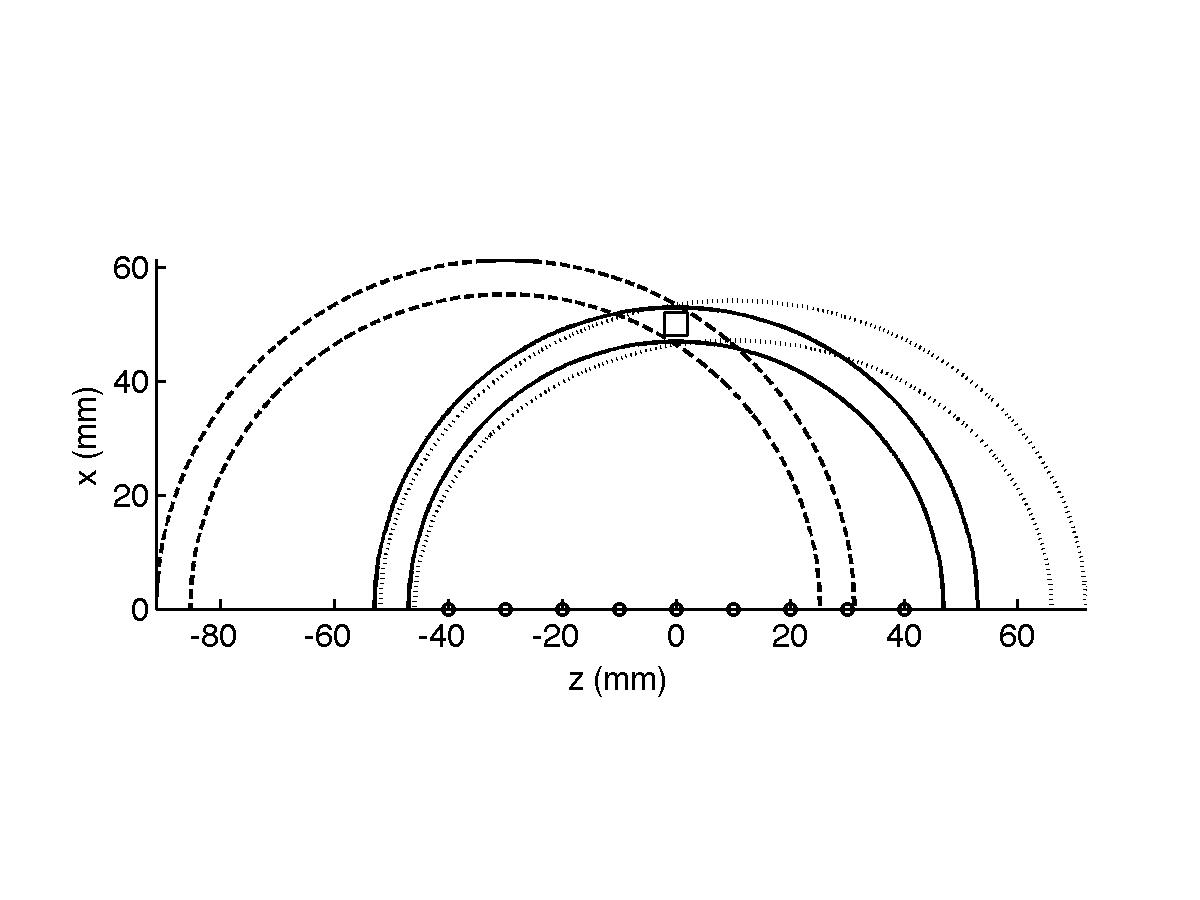}
		\caption{Elliptical rings showing the borders of the regions associated with the timing window for the corresponding element pairs of $(l,m)=(5,5)$ (solid), $(2,2)$ (dashed), and $(3,9)$ (dotted). The elements are marked with circles and have coordinates of $x = 0$, $z = \left\{-40,-30,...,40\right\}$ in millimeters. The region of interest (shown with a box) is located at the intersection of the annuli.} \label{fig:tdpic} 
	\end{figure}

The same window width is used for each transmitter-receiver pair where the center of the time-window is selected as the sum of the distances of the center of the region of interest to the transmitting and receiving elements divided by $c$. In spatial terms, we use only those scatterers that reside between the two ellipses with the two common focal points (location of transmitting and receiving elements) and radii $r_i$ and $r_o$, where $r_i = r_c - T c / 4$ and $r_o = r_c + T c / 4$ where $r_c$ is the sum of distances of the center of the region of interest to the two focal points. Figure~\ref{fig:tdpic} shows the spatial regions corresponding to the time-domain windows for element-pairs $(l,m)=(5,5)$ (middle element used in both transmission and reception), $(2,2)$ (second element from left used in both transmission and reception), and $(3,9)$ (the third element from left and the rightmost element used in transmission and reception, respectively), where the elements are numbered from 1 to 9 (from left to right).  Note that the the regions remain unaltered when the transmitting and receiving elements are swapped, i.e., element pair $(l,m)$ and $(m,l)$ use the same interval for time-windowing in the receiving element and have identical spatial regions in association.

\section*{Acknowledgements}
This work was supported in part by CenSSIS, the Center for Subsurface Sensing and Imaging Systems, under the Engineering Research Centers Program of the National Science Foundation (Award Number EEC-9986821) and by the NIH through R21 CA 123523.


\begin{thebibliography}{99}
\bibitem{bsk1} L.~Poissonnier, A.~Gelet, J.~Chapelon, R.~Bouvier, O.~Rouviere, C.~Pangaud, D.~Lyonnet, and J.~Dubernard, ``Results of transrectal focused ultrasound for the treatment of localized prostate cancer 120 patients with PSA or $+10$ ng/ml,'' \emph{Prog.~Urol.}, vol.~13, pg.~60--72, 2003.
\bibitem{bsk2} T.~A.~Gardner, M.~O.~Koch, A.~Shalhav, R.~Bihrle, R.~S.~Foster, C.~Steidle, I.~Grunberger, A.~S.~M.~Resnick, J.~Cochran, V.~Rao, and N.~T.~Sanghvi, ``Minimally invasive treatment of benign prostatic hyperplasia with high intensity focused ultrasound using the SonablateTM system: An updated report of phase III clinical studies conducted in the USA,'' \emph{Proc.~SPIE}, vol.~4609, pg.~107--114, 2002.
\bibitem{bsk3} N.~T.~Sanghvi, J.~Syrus, R.~S.~Foster, R.~Bihrle, R.~Casey, and T.~Uchida, ``Noninvasive surgery of prostate tissue by high intensity focused ultrasound: An updated report,Ó \emph{Proc.~SPIE}, vol.~3907, pg.~194--200, 2000.
\bibitem{bsk4} S.~Madersbacher, C.~Kratzik, and M.~Marberger, ``Prostatic tissue ablation by transrectal high intensity focused ultrasound: Histological impact and clinical application,'' \emph{Ultrason.~Sonochem.}, vol.~4, pg.~175--179, 1997.
\bibitem{bsk5} K.~Nakamura, S.~Baba, S.~Saito, M.~Tachibana, and M.~Murai, ``High-intensity focused ultrasound energy for benign prostatic hyperplasia: Clinical response at 6 months to treatment using sonablate 200,'' \emph{J.~Endourol}, vol.~11, pg.~197--201, 1997.
\bibitem{bsk6} E.~D. Mulligan, T.~H.~Lynch, D.~Mulvin, D.~Greene, J.~M.~Smith, and J.~M.~Fitzpatrick, ``High-intensity focused ultrasound in the treatment of benign prostatic hyperplasia,''  \emph{Br.~J.~Urol.}, vol.~79, pg.~177--180, 1997.
\bibitem{bsk7} C.~Chaussy, S.~Thuroff, F.~Lacoste, and A.~Gelet, ÒHIFU and prostate cancer: The European experience,Ó in \emph{Proceedings of the Second International Symposium on Therapeutic Ultrasound}, July 29--Aug. 1; Seattle, WA, 2002.
\bibitem{bsk8} J.~Kennedy, G.~T.~Haar, and D.~Cranstron, ÒHigh intensity focused ultrasound: Surgery of the future?,Ó \emph{Br.~J.~Radiol.}, vol.~76, pg.~590--599, 2003.
\bibitem{bsk9} F.~Wu, W.~Chen, J.~Bai, Z.~Zou, Z.~Wang, H.~Zhu, and Z.~Wang, ÒPathological changes in human malignant carcinoma treated with high-intensity focused ultrasound,Ó \emph{Ultrasound Med.~Biol.}, vol.~27, pg.~1099--1106, 2001.
\bibitem{bsk10} F.~Wu, Z.~Wang, W.~Chen, W.~Wang, Y.~Gui, M.~Zhang, G.~Zheng, Y.~Zhou, G.~Xu, M.~Li, C.~Zhang, H.~Ye, and R.~Feng, ÒExtracorporeal high intensity focused ultrasound ablation in the treatment of 1038 patients with solid carcinomas in China: An overview,Ó \emph{Ultrason.~Sonochem.}, vol.~11, pg.~149--154, 2004.
\bibitem{ebbini} D.~Liu, E.~S.~Ebbini, ``Real-time 2-D temperature imaging using ultrasound,'' \emph{Trans.~Biomed.~Eng.}, vol.~57, no.~1, pg.~12--16, 2009.
\bibitem{kaczkowski} P.~J.~Kaczkowski and A.~Anand, ``Monitoring high-intensity focused ultrasound (HIFU) therapy using radio frequency ultrasound backscatter to quantify heating (A),'' \emph{J.~Acoust.~Soc.~Am.}, vol.~118, no.~3, pp.~1882--1882, 2005.
\bibitem{hall} T.~L.~Hall, B.~Fowlkes, C.~A.~Cain, ``A real-time measure of cavitation induced tissue disruption by ultrasound imaging backscatter reduction,'' \emph{IEEE Trans.~Ultrason.~Ferroelectr.~Freq.~Control}, vol.~54, no.~3, pp.~569--575, 2007.
\bibitem{sanghvi} N.~T.~Sanghvi, F.~J.~Fry, R.~Bihrle, R.~S.~Foster, M.~H.~Phillips, J.~Syrus, A.~V.~Zaitsev, C.~W.~Hennige, ``Noninvasive surgery of prostate tissue by high-intensity focused ultrasound,''\emph{IEEE Trans.~Ultrason.~Ferroelectr.~Freq.~Control}, vol.~43, no.~6, pp.~1099--1110, 1996.
\bibitem{karbeyaz08} B.~U.~Karbeyaz, E.~L.~Miller, and R.~O.~Cleveland. ``Shape-based ultrasound tomography using a Born model with application to high intensity focused ultrasound therapy,'' \emph{Journal of the Acoustical Society of America}, vol.~123, pp.~2944--2956, May~2008.
\bibitem{vandongen} van~Dongen, K.~W.~A. and Wright, W.~M. D., ÒA full vectorial contrast source inversion scheme for three-dimensional acoustic imaging of both compressibility and density profiles,Ó \emph{J.~Acoust.~Soc.~Am.}, vol.~121, 1538--1549, 2007.
\bibitem{simon} C.~Simon, P.~VanBaren, and E.~S.~Ebbini, ``Two-dimensional temperature estimation using diagnostic ultrasound,'' \emph{IEEE Trans.~Ultrason.~Ferroelectr.~Freq.~Control}, vol.~45, pp.~1088--1099, 1998.
\bibitem{pernot} M.~Pernot, M.~Tanter, J.~Bercoff, K.~Waters, and M.~Fink, ÒTemperature estimation using ultrasonic spatial compound imaging,Ó \emph{IEEE Trans.~Ultrason.~Ferroelectr.~Freq.~Control}, vol.~51, pp.~606--615, 2004.
\bibitem{tyreus} P.~D.~Tyr\'eus and C.~Diederich, ``Two-dimensional acoustic attenuation mapping of high-temperature interstitial ultrasound lesions,'' \emph{Phys.~Med.~Biol.}, vol.~49, pg.~533--546, 2004.
\bibitem{devaney} A.~J.~Devaney, ``A filtered backpropagation algorithm for diffraction tomography,'' \emph{Ultrasonic Imaging}, vol.~4, pp.~336--350, 1982.
\bibitem{kak} A.~Kak and M.~Slaney, \emph{Principles of Computerized Tomographic Imaging}, SIAM Press, 1988.
\bibitem{EricPaper1} E.~L.~Miller, M.~Kilmer, and C.~Rappaport, ``A new shape-based method for object localization and characterization from scattered field data,'' \emph{IEEE Trans. Geoscience and Remote Sensing}, vol.~38, pp.~1682--1696, 2000.
\bibitem{EricPaper2} R.~Firoozabadi, E.~L.~Miller, C.~M.~Rappaport, and A.~W.~Morgenthaler, ``Subsurface sensing of buried objects under a randomly rough surface using scattered electromagnetic field data,Ó \emph{IEEE Transactions on Geoscience and Remote Sensing}, vol.~45, pp.~104--117, 2007.
\bibitem{burckhardt} C. B. Burckhardt, ``Speckle in ultrasound B-mode scans,'' \emph{IEEE Trans. Son. Ultrason.}, vol.~25, no.~1, pp.~1--6, 1978.
\bibitem{wagner} R.~F.~Wagner, S.~W.~Smith, J.~M.~Sandrik, and H.~Lopez, ``Statistics of Speckle in Ultrasound B-Scans,'' \emph{IEEE Trans. Son. Ultrason.}, vol.~30, no.~3, pp.~156--163, 1983.
\bibitem{rudin} L.~Rudin, S.~Osher, and E.~Fatemi, ``Nonlinear total variation based noise removal algorithms,'' \emph{Physica D}, vol.~60, no.~1, pp.~259--268, 1992.
\bibitem{zhou} Y.F. Zhou, S.G. Kargl, and J.H. Hwang ,   Producing uniform lesion pattern in HIFU ablation, 8th International Symposium on Therapeutic Ultrasound (Minneapolis, MN, USA) 91-95, 2008.
\bibitem{jensen1} J.~A.~Jensen, ``A model for the propagation and scattering of ultrasound in tissue,'' \emph{J.~Acoust.~Soc.~Am.}, vol.~89, pp.~182--191, 1991.
\bibitem{jensen2} J.~A.~Jensen and S.~Nikolov, ``Fast simulation of ultrasound images'', \emph{Proceedings the IEEE Ultrasonics Symposium}, Puerto Rico, Oct.~2000.
\bibitem{liuffc} Y.~Li and J.~A.~Zagzebski. ``A frequency domain model for generating B-mode images with array transducers,''  \emph{IEEE Trans.~Ultrason.~Ferroelectr.~Freq.~Control}, vol.~46, pp.~690--699, 1999.
\bibitem{chew} W.~C.~Chew, \emph{Waves and Fields in Inhomogeneous Media}, Van Nostrand Reinhold, New York, 1990.
\bibitem{szabo} T.~Szabo, \emph{Diagnostic Ultrasound Imaging: Inside Out}, Academic Press, 2004.
\bibitem{cramblitt} R.~M.~Cramblitt, K.~J.~Parker, ``Generation of non-Rayleigh speckle distributions using marked regularity models,'' {\em IEEE Trans.~Ultrason., Ferroelec., Freq.~Contr.}, vol.~46, no.~4, 1999.
\bibitem{scharf} L.~L.~Scharf, \emph{Statistical Signal Processing: Detection, Estimation, and Time Series Analysis}, Addison-Wesley, 1990.
\bibitem{karbeyazPhd}  B.~U.~Karbeyaz, ``Modeling and shape based inversion for frequency domain ultrasonic monitoring of cancer treatment,'' PhD thesis, Northeastern University, 2005.
\bibitem{roys1} J.~P.~Royston, ``An extension of Shapiro and Wilk's W-test for normality to large samples,'' \emph{Applied Statistics}, vol.~31, no.~2, pp.~115--124, 1982.
\bibitem{roysMatlab} A.~Trujillo-Ortiz, R.~Hernandez-Walls, K.~Barba-Rojo, and L.~Cupul-Magana. (2007). Roystest:Royston's Multivariate Normality Test. A MATLAB file. [WWW document]. URL http://www.mathworks.com/matlabcentral/fileexchange/17811-roystest
\bibitem{vauhkonen} M.~Vauhkonen, D.~Vad\'asz, P.~A.~Karjalainen, E.~Somersalo, and J.~P.~Kaipio ``Tikhonov regularization and prior information in electrical impedance tomography'' {\em IEEE Trans. Med. Imag.}, vol.~17, pp.~285--293, 1998.
\bibitem{candesNoisy} E.~J.~Cand{\`e}s, J.~Romberg, and T.~Tao, ``Stable signal recovery from incomplete and inaccurate measurements,'' \emph{Comm.~Pure Appl.~Math.}, vol.~59, pp.~1207--1223, 2005.
\bibitem{vogel} C.~R.~Vogel. Computational methods for inverse problems. Philadelphia, PA, SIAM, 2002.
\bibitem{wen} Y.-W.~Wen and A.~M.~Yip, ``Adaptive parameter selection for total variation image deconvolution,'' \emph{Numer. Math. Theor. Meth. Appl.}, vol.~2, no.~4, pp.~427-438, 2009.
\bibitem{strong} D.~M.~Strong, J.-F.~Aujol, and T.~F.~Chan, ``Scale recognition, regularization parameter selection, and Meyer's G norm in total variation regularization,'' \emph{Multiscale Model.~Simul.}, vol.~5, no.~1, pp.~273-303, 2006.
\bibitem{morozov} V.~Morozov, ``On the solution of functional equations by the method of regularization,''  \emph{Soviet Math. Doklady}, vol.~7, pp.~414--417, 1966.
%
\bibitem{saad} Y.~Saad and M.~H.~Schultz, ``GMRES: A generalized minimal residual algorithm for solving nonsymmetric linear systems,'' \emph{SIAM J.~Sci.~Stat.~Comput.}, vol.~7, no.~3, pp.~856--869, 1986.
\bibitem{longley} L.~A.~Longley and W.~D.~O'Brien, Jr., ``Ultrasonic heating distribution in lossy cylinders and spheres,'' \emph{IEEE Trans. Sonics and Ultrasonics}, vol.~SU-29, no.~2, 1982.
 \bibitem{seip} R.~Seip, J.~Tavakkoli, R.~F.~Carlson, A.~Wunderlich, N.~T.~Sanghvi, K.~A.~Dines, T.~A.~Gardner, ``High-intensity focused ultrasound (HIFU) multiple lesion imaging: comparison of detection algorithms for real-time treatment control,'' \emph{Proceedings the IEEE Ultrasonics Symposium}, vol.~2, pg.~1427--1430, 2002.
 \bibitem{ziadloo} A.~Ziadloo, V.~Shahram, ``Real-time 3D image-guided HIFU therapy,'' Engineering in Medicine and Biology Society, 2008. EMBS 2008. 30th Annual International Conference of the IEEE, pg.~4459--4462, 2008.
 %
\bibitem{kitap}J.~A.~Zagzebski. Essentials of ultrasound physics. Elsevier, 1996.
\bibitem{cvx} M.~Grant and S.~Boyd. CVX: Matlab software for disciplined convex programming (web page and software). http://stanford.edu/$\sim$boyd/cvx, August 2008.
\bibitem{dcp} M.~Grant and S.~Boyd. Graph implementations for nonsmooth convex programs, Recent Advances in Learning and Control (a tribute to M.~Vidyasagar), V.~Blondel, S.~Boyd, and H.~Kimura, editors, pages 95--110, Lecture Notes in Control and Information Sciences, Springer, 2008. http://stanford.edu/$\sim$boyd/graph\_dcp.html.
\bibitem{mattingley} J.~Mattingley and S.~Boyd, ``Real-time convex optimization in signal processing,'' \emph{IEEE Signal Processing Magazine}, vol.~27, pg.~50--61, 2010.
\end{thebibliography}
\end{document}